\documentclass[journal]{IEEEtran}
\ifCLASSINFOpdf
\else
\fi

\usepackage{cite}
\usepackage{graphicx}
\usepackage{amsmath,amssymb} % define this before the line numbering.
\usepackage{amsfonts}
\usepackage{color}
\usepackage{units}
\usepackage{cases}
\usepackage{floatrow}
\usepackage{float}
\usepackage{array}
\usepackage{mathrsfs}
\usepackage{mathtools}
\usepackage{stfloats}
\usepackage{algorithm}
\usepackage{algorithmic}
\usepackage{multirow}
\usepackage{float}
\usepackage{tabularx}
\usepackage{nicefrac}
\usepackage{amssymb}
\usepackage{url}

\newcommand{\Lagr}{\mathcal{L}}

\newcolumntype{L}[1]{>{\raggedright\arraybackslash}p{#1}}
\newcolumntype{C}[1]{>{\centering\arraybackslash}p{#1}}
\newcolumntype{R}[1]{>{\raggedleft\arraybackslash}p{#1}}

\begin{document}
\title{Adaptive Image Sampling using Deep Learning and its Application on X-Ray Fluorescence Image Reconstruction}

\author{Qiqin~Dai,	
		Henry Chopp,
        Emeline Pouyet,
        Oliver Cossairt,
        Marc Walton,
        and~Aggelos~K.~Katsaggelos,~\IEEEmembership{Fellow,~IEEE}% <-this % stops a space
\thanks{Q. Dai, H. Chopp, O. Cossairt and A. K. Katsaggelos are with the Department
of Electrical Engineering and Computer Science, Northwestern University, Evanston, IL, 60208 USA. E. Pouyet and M. Walton are with the Northwestern University / Art Institute of Chicago Center for Scientific Studies in the Arts (NU-ACCESS), Evanston, IL, 60208 USA.}% <-this % stops a space
\thanks{Manuscript received September 15, 2018.}}


\markboth{Journal of \LaTeX\ Class Files,~Vol.~11, No.~4, December~2012}%
{Shell \MakeLowercase{\textit{et al.}}: Bare Demo of IEEEtran.cls for Journals}



\maketitle

\begin{abstract}
This paper presents an adaptive image sampling algorithm based on Deep Learning (DL). It consists of an adaptive sampling mask generation network which is jointly trained with an image inpainting network. The sampling rate is controlled by the mask generation network, and a binarization strategy is investigated to make the sampling mask binary. In addition to the image sampling and reconstruction process, we show how it can be extended and used to speed up raster scanning such as the X-Ray fluorescence (XRF) image scanning process. Recently XRF laboratory-based systems have evolved into lightweight and portable instruments thanks to technological advancements in both X-Ray generation and detection. However, the scanning time of an XRF image is usually long due to the long exposure requirements (e.g., $100 \mu s-1ms$ per point). We propose an XRF image inpainting approach to address the long scanning times, thus speeding up the scanning process, while being able to reconstruct a high quality XRF image. The proposed adaptive image sampling algorithm is applied to the RGB image of the scanning target to generate the sampling mask. The XRF scanner is then driven according to the sampling mask to scan a subset of the total image pixels. Finally, we inpaint the scanned XRF image by fusing the RGB image to reconstruct the full scan XRF image. The experiments show that the proposed adaptive sampling algorithm is able to effectively sample the image and achieve a better reconstruction accuracy than that of existing methods.

\end{abstract}

\begin{IEEEkeywords}
Adaptive sampling, convolutional neural network, X-Ray fluorescence, inpainting
\end{IEEEkeywords}

% \fi
\IEEEpeerreviewmaketitle

\section{Introduction}
\label{sec:introduction}

\IEEEPARstart{W}{ith} the increasing demand for multimedia content, there has been more and more interest in visual data acquisition. Many visual data, such as Lidar depth map, scanning probe microscopy (SPM) image, XRF image, etc, is acquired by the time consuming raster scan process. Thus sampling techniques need to be investigated to speed up the acquisition process. Compressed sensing (CS) has shown that it is possible to acquire and reconstruct natural images under the Nyquist sampling rates \cite{candes2006robust, donoho2006compressed}. Rather than full image acquisition followed by compression, CS combines sensing and compression into one step, and has the advantages of faster acquisition time, smaller power consumption, and lighter data throughput. Adaptive image sampling is a sub-problem of CS that aims for a sparse representation of signals in the image domain. In this paper, we present a novel adaptive image sampling algorithm based on Deep Learning and show its application to RGB image sampling and recovery. We also applied the proposed adaptive sampling technique to  speed up the raster scan process of XRF imaging, based on the correlation between RGB and XRF signals.

Irregular sampling techniques have long been studied in the image processing and computer graphics fields to achieve compact representation of images. Such irregular sampling techniques, such as stochastic sampling~\cite{cook1986stochastic}, may have better anti-aliasing performance compared to uniform sampling intervals if frequencies greater than the Nyquist limit are present. Further performance improvement can be obtained if the sampling distribution is not only irregular but also adaptive to the signal itself. The limited samples should be concentrated in parts of the image rich in detail, so as to simulate human vision~\cite{soumekh1998multiresolution}. Several works have been reported in the literature on adaptive sampling techniques. An early significant work in this direction is made by Eldar \textit{et al.}~\cite{eldar1997farthest}. A farthest point strategy is proposed which permits progressive and adaptive sampling of an image. Later, Rajesh \textit{et al.}~\cite{rajesh2007fast} proposed a progressive image sampling technique inspired by the lifting scheme of wavelet generation. A similar method is developed by Demaret \textit{et al.}~\cite{demaret2006image} by utilizing an adaptive thinning algorithm. Ramponi \textit{et al.}~\cite{ramponi2001adaptive} developed an irregular sampling method based on a measure of the local sample skewness. Lin \textit{et al.}~\cite{lin2015generalized} viewed grey scale images as manifolds with density and sampled them according to the generalized Ricci curvature. Liu \textit{et al.}~\cite{liu2014kernel} proposed an adaptive progressive image acquisition algorithm based on kernel construction. {Recently, Taimori \textit{et al.}~\cite{taimori2018adaptive} investigated adaptive image sampling approaches based on the space-frequency-gradient information content of image patches.}

Most of these irregular sampling and adaptive sampling techniques~\cite{cook1986stochastic, eldar1997farthest, rajesh2007fast, demaret2006image, ramponi2001adaptive, lin2015generalized, liu2014kernel, taimori2018adaptive} need their own specific reconstruction algorithm to reconstruct the fully sampled signal. Furthermore, all these sampling techniques are model-based approaches relying on predefined priors, and according to our knowledge, no work has been done on utilizing machine learning techniques to design the adaptive sampling mask.

Inspired by the recent successes of convolutional neural networks (CNNs)~\cite{krizhevsky2012imagenet, szegedy2015going, xie2019automated, xie2019convolutional} in high level computer vision tasks, deep neural networks (DNNs) emerged in addressing low level computer vision tasks as well~\cite{vincent2008extracting, dong2014learning, kappeler2016video, iliadis2018deep, iliadis2016deepbinarymask, pathak2016context, yeh2016semantic, gao2016one}. For the task of image inpainting, Pathak \textit{et al.}~\cite{pathak2016context} presented an auto-encoder to perform context-based image inpainting. The inpainting performance is improved by introducing perceptual loss~\cite{yeh2016semantic} and on-demand learning~\cite{gao2016one}. Iliadis \textit{et al.}~\cite{iliadis2018deep} utilized a deep-fully-connected network for video compressive sensing while also learning an optimal binary sampling mask~\cite{iliadis2016deepbinarymask}. However, the learned optimal binary sampling mask is not adaptive to the input video signals. According to our knowledge, no work has been made on generating the adaptive binary sampling mask for the image inpainting problem using deep learning.

With the proposed adaptive sampling algorithm, we can efficiently sample points based on the local structure of the input images. Besides the application of Progressive Image Transmission (PIT)~\cite{tzou1987progressive}, compressed image sampling, image coding~\cite{deng2012robust, liu2014joint, zhang2016bi, chen2018efficient, xie2018double}, etc., we also show that the proposed adaptive sampling algorithm can speed up many raster scan processes such as XRF imaging. In detail, the binary sampling mask can be obtained using one modality (RGB image for example) of the target, and then applied on the raster scan process for another modality (XRF image for example). A detailed introduction to XRF imaging is in Section~\ref{sec:xrfimaging}.

The contribution of this paper lies in the following aspects:

\begin{itemize}
  \item We proposed an effective way to binarize and control the sparseness of the output of a CNN.
  \item We proposed an efficient network structure to generate the binary sampling mask and showed its advantages over other state-of-the-art adaptive sampling algorithms.
  \item We proposed an adaptive sampling framework that can be applied to speed up many raster scan processes.
  \item We proposed a fusion-based image reconstruction algorithm to restore the fully sampled XRF image.
  \item We experimentally showed that the benefits of the adaptive sampling and the fusion-based inpainting algorithm are additive.
\end{itemize}

This paper is organized as follows: Section~\ref{sec:xrfimaging} introduces XRF imaging with adaptive sampling. We illustrate the adaptive sampling mask design in Section~\ref{sec:cnnadapsampling}. We describe the XRF image inpainting problem in Section~\ref{sec:spatialspectralXRFinpainting}. In Section~\ref{sec:experimenntalresults}, we provide the experimental results with both synthetic data and real data to evaluate the effectiveness of the proposed approach. The paper is concluded in Section~\ref{sec:conclusion}.

\begin{figure}[b]
\begin{tabular}{c c }
\includegraphics[width=0.4\textwidth]{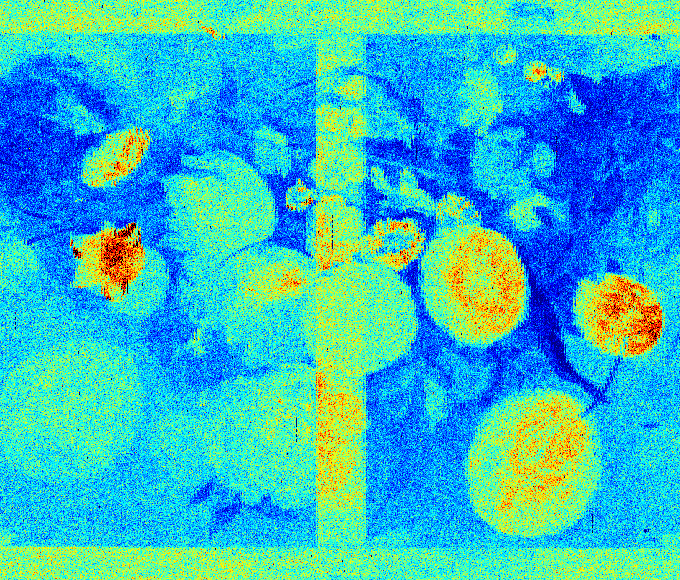}	    &
\includegraphics[width=0.4\textwidth]{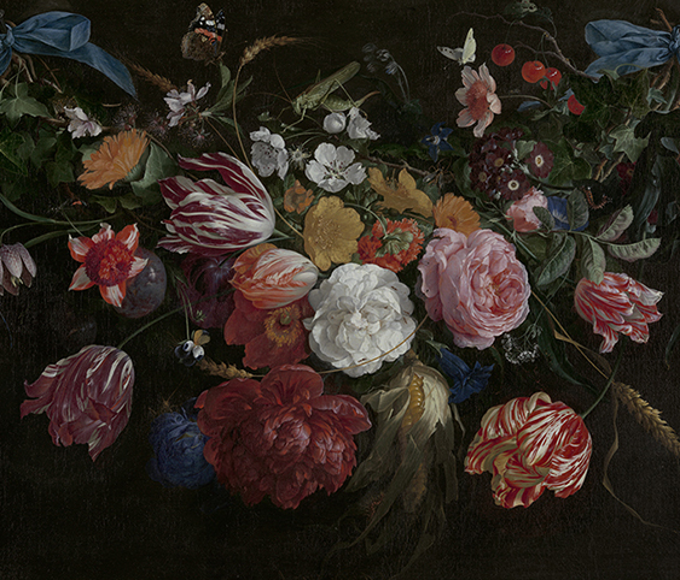}	\\ \centering{(a)}   &    \centering{(b)}
\end{tabular}
\caption{(a) XRF map showing the distribution of $Pb\ L\eta$ XRF emission line (sum of channel \#582 - 602) of the ``Bloemen en insecten'' (ca 1645), by Jan Davidsz. de Heem, in the collection of Koninklijk Museum voor Schone Kunsten (KMKSA) Antwerp and (b) the HR RGB image.}
\label{fig:xrf_example}
\end{figure}

% I wish you the best of success.


% \hfill December 27, 2012

\section{XRF Imaging Using Adaptive Sampling}
\label{sec:xrfimaging}

During the past few years, XRF laboratory-based systems have evolved to lightweight and portable instruments thanks to technological advancements in both X-Ray generation and detection. Spatially resolved elemental information can be provided by scanning the surface of the sample with a focused or collimated X-ray beam of (sub) millimeter dimensions and analyzing the emitted fluorescence radiation in a nondestructive in-situ fashion entitled Macro X-Ray Fluorescence (MA-XRF). The new generations of XRF spectrometers are used in the Cultural Heritage field to study the manufacture, provenance, authenticity, etc. of works of art. Because of their fast noninvasive set up, we are able to study large, fragile, and location-inaccessible art objects and archaeological collections. In particular, XRF has been used extensively to investigate historical paintings by capturing the elemental distribution images of their complex layered structure. This method reveals the painting history from the artist creation to restoration processes~\cite{alfeld2013mobile, anitha2013restoration}.

As with other imaging techniques, high spatial resolution and high quality spectra are desirable for XRF scanning systems; however, the acquisition time is usually limited, resulting in a compromise between dwell time, spatial resolution, and desired image quality. In the case of scanning large scale mappings, a choice may be made to reduce the dwell time and increase the step size, resulting in noisy XRF spectra and low spatial resolution XRF images.

An example of an XRF scan is shown in Figure~\ref{fig:xrf_example} (a). Channel $\#582 - 602$ corresponding to the $Pb\ L\eta$ XRF emission line was extracted from a scan of Jan Davidsz. de Heem's ``Bloemen en insecten'' painted in $1645$ (housed at Koninklijk Museum voor Schone Kunsten (KMKSA) Antwerp). The image is color coded for better visibility. This XRF image was collected by a home-built XRF spectrometer (courtesy of Prof. Koen Janssens) with 2048 channels in spectrum and spatial resolution $680 \times 580$ pixels. This scan has a relatively short dwell time, resulting in low Signal-to-Noise Ratio (SNR), yet it still took $18$ hours to acquire it. Faster scanning speed will be desirable for promoting the popularity of the XRF scanning technique, since the slow acquisition process impedes the use of XRF scanning instruments as high resolution widefield imaging devices. The RGB image of the painting of resolution $680 \times 580$ pixels is shown in Figure~\ref{fig:xrf_example} (b).

\begin{figure}[t]
\begin{tabular}{c c }
\includegraphics[width=0.4\textwidth]{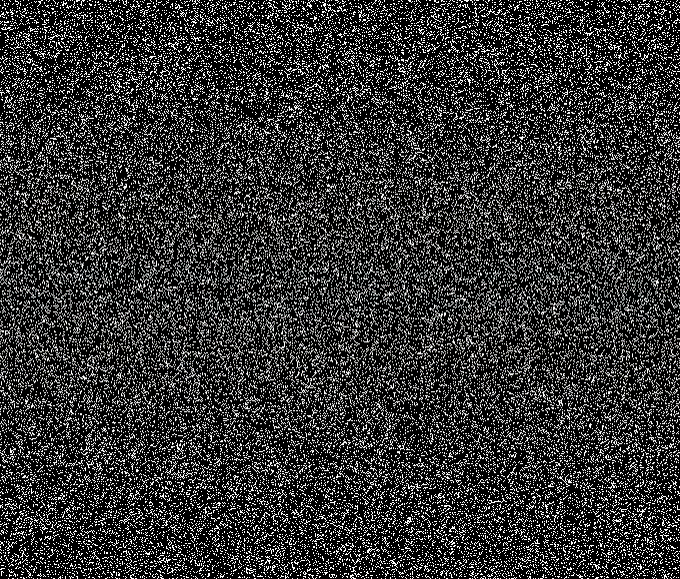}	    &
\includegraphics[width=0.4\textwidth]{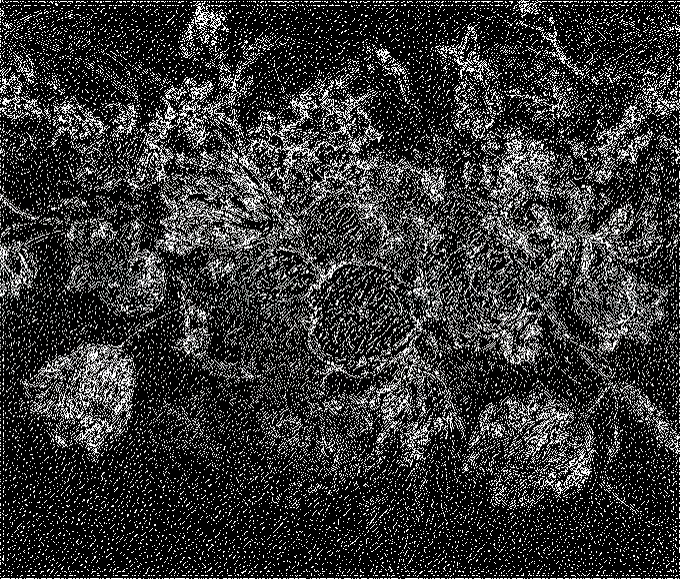}	\\ \centering{(a)}   &    \centering{(b)}
\end{tabular}
\caption{(a) Random binary sampling mask that skips 80\% of pixels and (b) Adaptive binary sampling mask that skips 80\% of pixels based on the input RGB images in Fig~\ref{fig:xrf_example} (b). }
\label{fig:deheem_sampling_example}
\end{figure}

\begin{figure*}[h]
\centering
\includegraphics[width=0.45\linewidth]{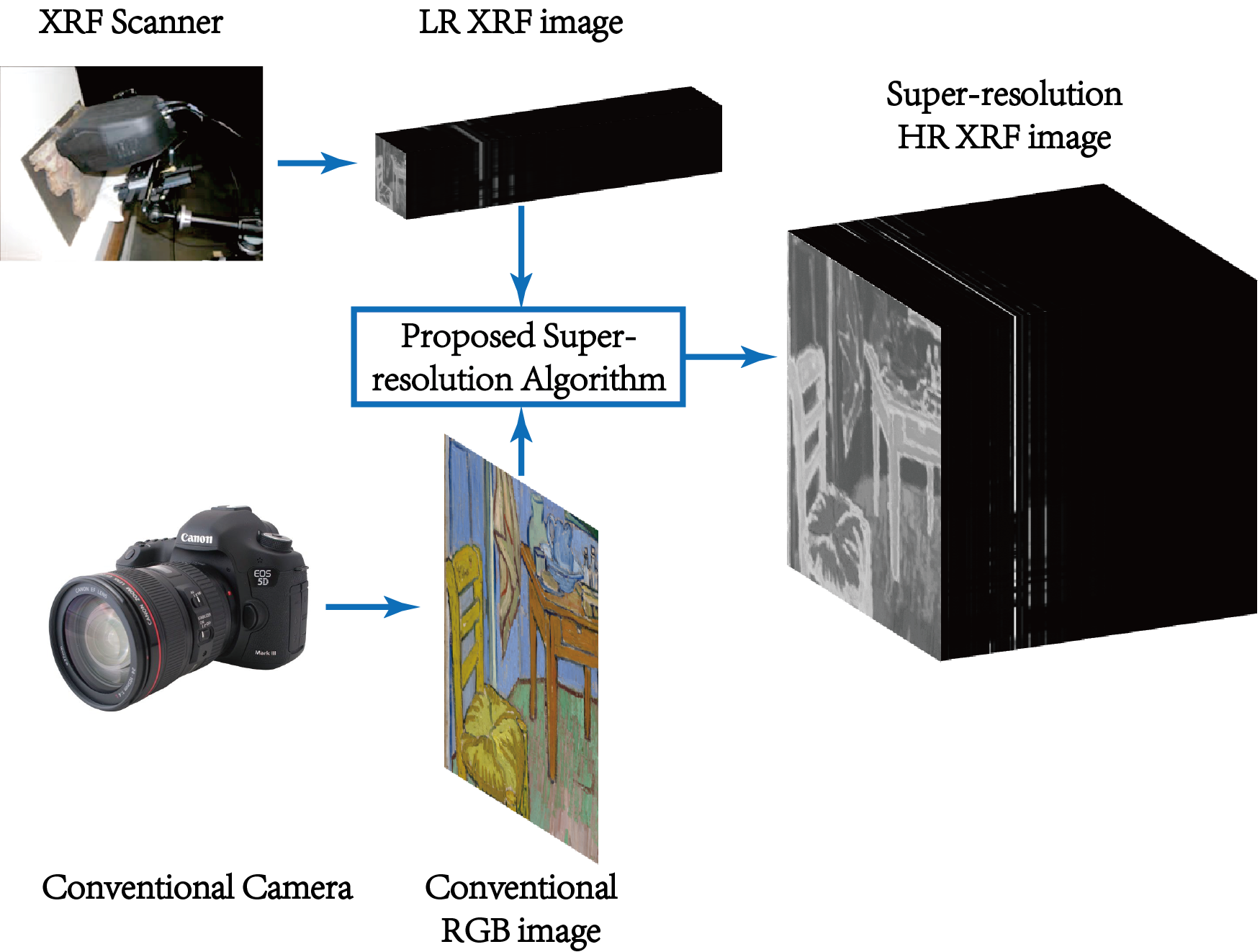}
\caption{The proposed pipeline for the XRF image inpainting utilizing an adaptive sampling mask. The binary adaptive sampling mask is generated based on the RGB image of the scan target. Then, the XRF scanner sampled the target object based on the binary sampling mask. Finally, the subsampled XRF image and the RGB image are fused to reconstruct the fully sampled XRF image.}
\label{fig:XRFInpaintingDiagram}
\end{figure*}

\begin{figure}[h]
\centering
\includegraphics[width=0.75\linewidth]{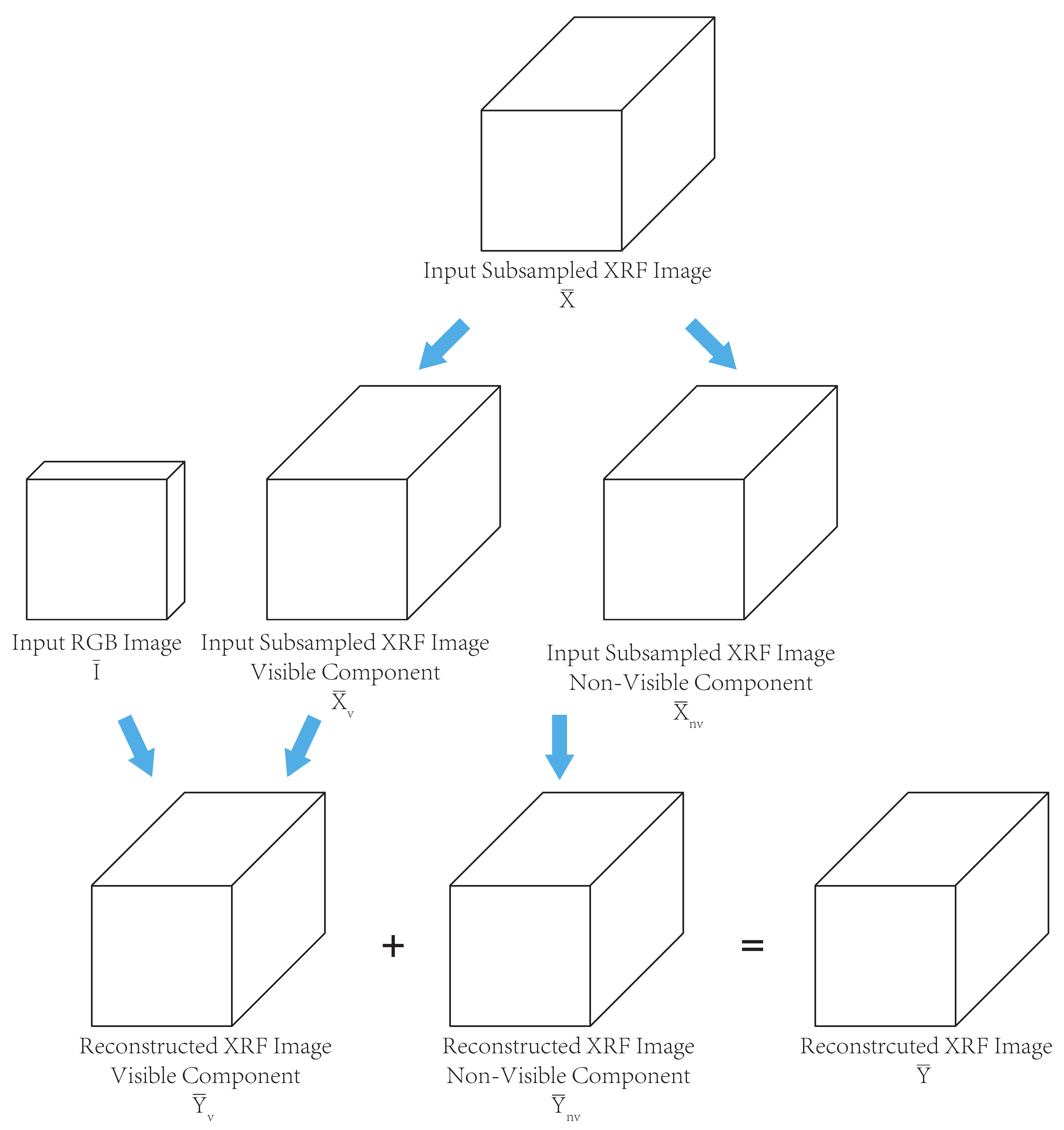}
\caption{Proposed pipeline of XRF image inpainting. The visible component of the input subsampled XRF image is fused with the input RGB image to obtain the visible component of the reconstructed XRF image. The non-visible component of the input XRF image is super-resolved to obtain the non-visible component of the reconstructed XRF image. The reconstructed visible and non-visible component of the output XRF image are combined to obtain the final output.}
\label{fig:XRFInpaintingVNVPipeline}
\end{figure}

Image inpainting~\cite{bertalmio2000image, criminisi2004region, bertalmio2003simultaneous} is the process of recovering missing pixels in images. The XRF images are acquired through a raster scan process. We could therefore speed up the scanning process by skipping pixels and then utilizing an image inpainting technique to reconstruct the missing pixels. If we are to skip $80\%$ of the pixels during acquisition (a $5$x speedup), we could use a random sampling mask (shown in Figure~\ref{fig:deheem_sampling_example} (a)), or we could design one utilizing the available RGB image (shown in Figure~\ref{fig:deheem_sampling_example} (b)). The idea of the adaptive binary sampling mask is based on the assumption that the XRF image is highly correlated with the RGB image. We would like to allocate more pixels to the informative parts of the image, such as high frequency textures, sharp edges, and high contrast details, and spend fewer pixels on the uninformative parts of the image.

With the proposed adaptive sampling algorithm, we propose an image inpainting approach to speed up the acquisition process of the XRF image with the aid of a conventional RGB image, as shown in Figure~\ref{fig:XRFInpaintingDiagram}. The proposed XRF image inpainting algorithm can also be applied to spectral images obtained by any other raster scanning processes, such as Scanning Electron Microscope (SEM), Energy Dispersive Spectroscopy (EDS), and Wavelength Dispersive Spectroscopy (WDS). First, the RGB image of the scanning target is utilized to generate the adaptive sampling mask. Then, the XRF scanner will scan the corresponding pixels according to the binary sampling mask. The speedup in acquisition is achieved since many pixels will be skipped. Finally, the subsampled XRF image is fused with the conventional RGB image to reconstruct the full scan XRF image, utilizing an image inpainting algorithm. For the fusion-based XRF image inpainting algorithm, similarly to our previous super-resolution (SR) approach~\cite{dai2016x, dai2017x}, we model the spectrum of each pixel using a linear mixing model~\cite{manolakis2001hyperspectral}. Because the hidden part of the painting is not visible in the conventional RGB image, but it can be captured in the XRF image~\cite{alfeld2013revealing}, there is no direct one-to-one mapping between the visible RGB spectrum and the XRF spectrum. We model the XRF signal as a combination of the visible signal (on the surface) and the non-visible signal (hidden under the surface), as shown in Figure~\ref{fig:XRFInpaintingVNVPipeline}. We emphasize that while our framework is general enough to handle separation of visible and hidden layers, it easily handles the case of fully visible layers. To inpaint the visible component XRF signal, we follow an approach similar to the one applied to hyper-spectral image SR~\cite{akhtar2014sparse, lanaras2015hyperspectral, dong2016hyperspectral}. A coupled XRF-RGB dictionary pair is learned to explore the correlation between XRF and RGB signals. For the non-visible part, we inpaint its missing pixels using a standard total variation regularizer. Finally, the reconstructed visible and non-visible XRF signals are combined to obtain the final XRF reconstruction result. The input subsampled XRF image is not explicitly separated into visible and non-visible parts in advance. Instead, the whole inpainting problem is formulated as an optimization problem. By alternately optimizing over the coupled XRF-RGB dictionary and the visible/non-visible fully sampled coefficient maps, the fidelity of the estimated fully sampled output to both the subsampled XRF and RGB input signals is improved, thus resulting in a better inpainting output. Real experiments show the effectiveness of our proposed method in terms of both reconstruction error and visual quality of the inpainting result.

While there is a large body of work on inpainting conventional RGB images~\cite{bertalmio2000image, criminisi2004region, bertalmio2003simultaneous, esedoglu2002digital, shen2002mathematical, zhou2012nonparametric, pathak2016context, yeh2016semantic, gao2016one}, very little work has appeared in the literature on inpainting XRF images~\cite{bertalmio2000image}, and there is no work on fusing a conventional RGB image during the inpainting process. {Some learning based approaches~\cite{song2018multi, shivakumar2019dfusenet} have been proposed to perform multi-modality image inpainting. However, due to the limited amount of high quality XRF image training data, they can not be applied yet.} XRF image inpainting poses a particular challenge because the acquired spectrum signal usually has low SNR. In addition, the correlation among spectral channels needs to be preserved for the inpainted pixels. In our previous work on spatial-spectral representation for XRF image super-resolution~\cite{dai2017x}, the spatial resolution of the visible component XRF signal is increased by fusing an HR conventional RGB image while the spatial resolution of the non-visible part is increased by using a standard total variation regularizer~\cite{babacan2008total, marquina2008image}. Here we propose an XRF image inpainting algorithm by fusing an HR conventional RGB image, which can be regarded as an extension of our previous XRF SR approach.

\section{Adaptive Sampling Mask Generation utilizing Convolutional Neural Network}
\label{sec:cnnadapsampling}

In this section, we present our proposed adaptive sampling mask generation using a CNN. In other words, we describe the details of the ``Sampling Mask Generation'' block in Figure~\ref{fig:XRFInpaintingDiagram}. We first formulate the problem of adaptive sampling mask design, followed by the presentation of the overall network architecture consisting of both the inpainting network and the mask generation network.

\subsection{Problem Formulation}
\label{sec:cnnmaskproblemformulation}

\begin{figure*}[t]
\centering
\includegraphics[width=0.7\linewidth]{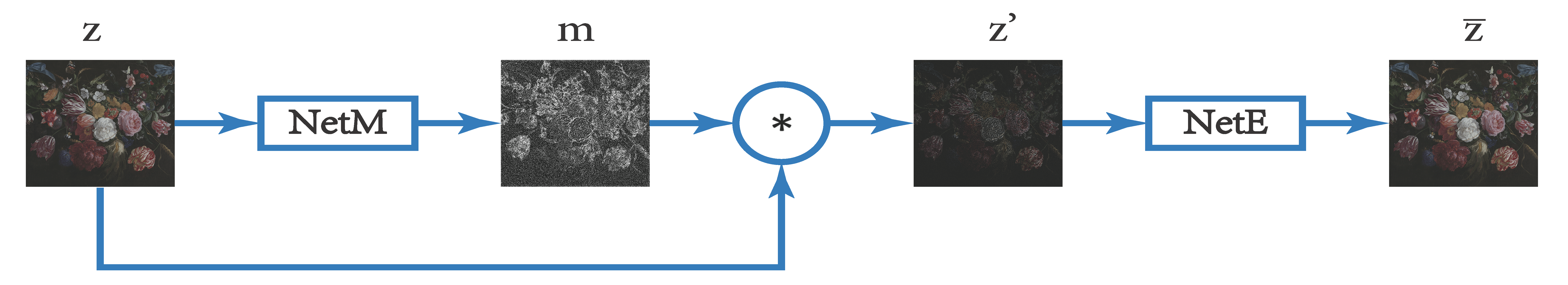}
\caption{Pipeline for adaptive sampling mask generation utilizing CNNs.}
\label{fig:pipelineadaptive}
\end{figure*}

\begin{figure*}[h]
\centering
\includegraphics[width=0.75\linewidth]{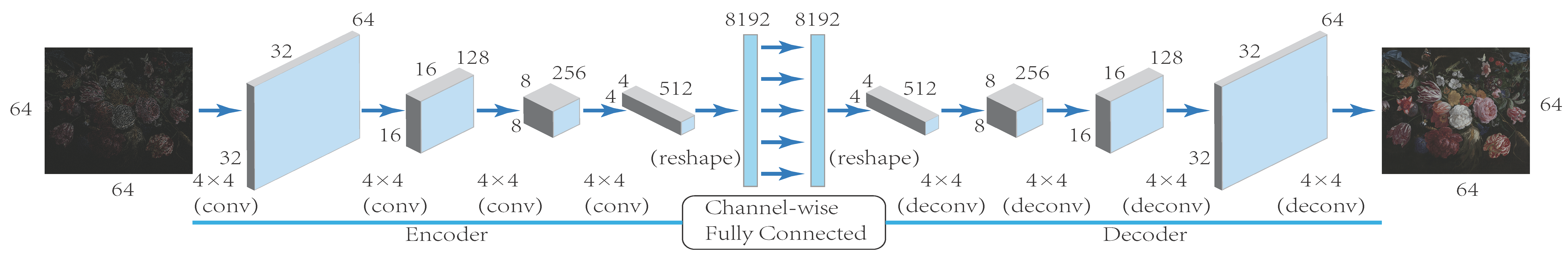}
\caption{Network architecture for the image inpainting network ($NetE$). The inpainting framework is an autoencoder style network with the encoder and decoder connected by a channel-wise fully-connected layer. }
\label{fig:inpaintingNet}
\end{figure*}

As shown in Figure~\ref{fig:pipelineadaptive}, we denote by $z$ an input original image. Our mask generation network $NetM$ produces a binary sampling mask $m = NetM(z, c)$,  where $c \in [0\ 1]$ is the predefined sampling percentage. The entries of $m$ are equal to $1$ for the sampled and $0$ otherwise. The corrupted image $z^{\prime}$ is obtained by

\begin{equation}
\label{e:e1}
z^{\prime} = z \odot m = z \odot NetM(z, c),
\end{equation} where $\odot$ is the element-wise product operation. The reconstructed image $\bar{z}$ is obtained by the inpainting network $NetE$,

\begin{equation}
\label{e:e2}
\bar{z} = NetE(z^{\prime}) = NetE(z \odot NetM(z, c)).
\end{equation}

The overall pipeline is shown in Figure~\ref{fig:pipelineadaptive}. We could regard the whole pipeline (Equation~\ref{e:e2}) as one network with input $z$ and output $\bar{z}$ and perform an end to end training. If we simultaneously optimize the mask generation network $NetM$ and the inpainting network $NetE$ according to the following loss function,

\begin{equation}
\label{e:e3}
\Lagr(z) = \| z - \bar{z} \|_2 = \| z - NetE(z \odot NetM(z, c)) \|_2,
\end{equation}
$NetM$ will perform an optimized adaptive sampling strategy according to the input image, and $NetE$ will perform optimized image inpainting. After the mask has been generated by the network $NetM$, we can replace the inpainting network $NetE$ with other image inpainting algorithms. The detailed network architecture of $NetE$ and $NetM$ are discussed in the following two subsections~\ref{sec:inpanitarchitecture} and \ref{sec:maskarchitecture}, respectively. {The detailed training procedure of $NetM$ and $NetE$ will be discussed in Section~\ref{sec:implementationDetails}}

\subsection{Deep Learning Network Architecture for Inpainting Network}
\label{sec:inpanitarchitecture}

The network architecture in~\cite{gao2016one} is used for the inpainting network, as shown in Figure~\ref{fig:inpaintingNet}. The network is an encoder-decoder pipeline. The encoder takes a corrupted image $z^{\prime}$ of size $64 \times 64$ as input and encodes it in the latent feature space. The decoder takes the feature representation and outputs the restored image $\bar{z}=NetE(z^{\prime})$. The encoder and decoder are connected through a channel-wise fully-connected layer. For the encoder, four convolutional layers are utilized. A batch normalization layer~\cite{ioffe2015batch} is placed after each convolutional layer to accelerate the training speed and stabilize the learning process. The Leaky Rectified Linear Unit (LeakyReLU) activation~\cite{maas2013rectifier, xu2015empirical} is used in all layers in the encoder.

The convolutional layers in the encoder only connect all the feature maps together, but there are no direct connections among different locations within each specific feature map. Fully-connected layers are then applied to handle this information propagation. To reduce the number of parameters in the fully connected layers, a channel-wise fully-connected layer is used to connect the encoder and decoder, as in~\cite{pathak2016context}. The channel-wise fully connected layer is designed to only propagate information within activations of each feature map. This significantly reduces the number of parameters in the network and accelerates the training process.

The decoder consists of four deconvolutional layers~\cite{long2015fully, dosovitskiy2015learning, zeiler2014visualizing}, each of which is followed by a ReLU activation except for the output layer. The tanh function is used in the output layer to restrict the pixel range of the output image. The series of up-convolutions and nonlinearities perform a nonlinear weighted upsampling of the feature produced by the encoder and generates an inpainted image of the target size ($64 \times 64$).

\subsection{Deep Learning Network Architecture for the Mask Generation Network}
\label{sec:maskarchitecture}

\begin{figure*}[b]
\centering
\includegraphics[width=0.75\linewidth]{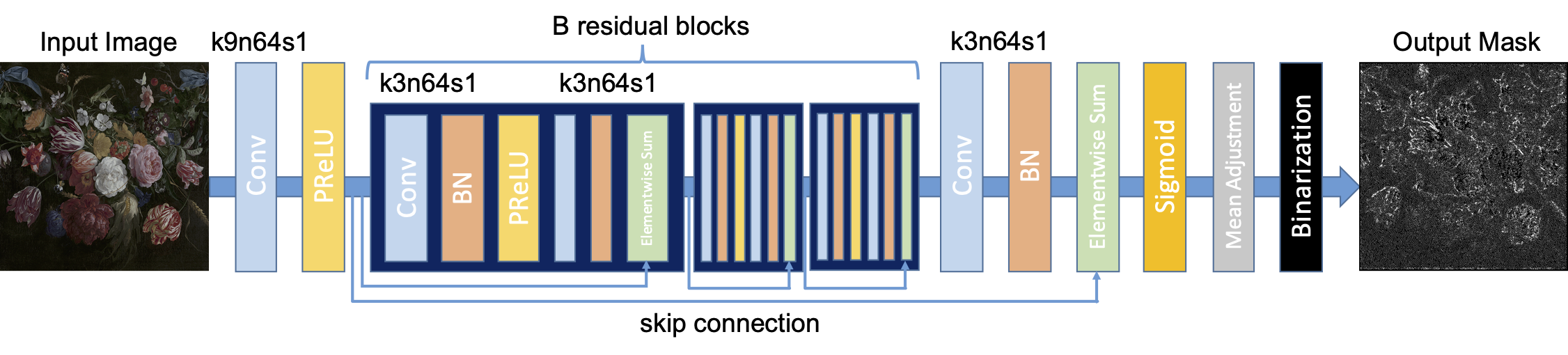}
\caption{Network architecture for the mask generation network ($NetM$). Residual blocks with skip connections are used in the binary sampling mask network. }
\label{fig:maskNet}
\end{figure*}

According to our knowledge, no prior work has been reported on generating the adaptive binary sampling mask utilizing CNNs. The desired mask generation network $NetM$ should satisfy the following criteria:

\begin{itemize}
  \item The output image should have the same spatial resolution as the input image.
  \item The network architecture should be fully convolutional to handle arbitrary input sizes.
  \item The output image should be binary.
  \item The output image should have a certain percentage $c$ of 1's.
\end{itemize}

Inspired by~\cite{johnson2016perceptual}, a network architecture with residual blocks is applied here, as shown in Figure~\ref{fig:maskNet}. The network $NetM$ consists of \textit{B} residual blocks with identical layout. Following the method used in~\cite{sam2016torch}, we use two convolutional layers with small $3\times3$ kernels and $64$ feature maps followed by batch-normalization layers~\cite{ioffe2015batch} and ParametricReLU~\cite{he2015delving} as the activation function. The network $NetM$ is fully convolutional to handle an arbitrary input size. To keep the spatial dimensions at each layer the same, the images are padded with zeros. At the end of each residual block is a final elementwise sum layer, followed by a Sigmoid activation layer. Let us denote by $L_{ij}$ the $(i,j)^{th}$ element of $L$, which is the output of the Sigmoid activation layer, and mapped to the range $[0\ 1]$. The Mean Adjustment layer $F$ is defined as

\begin{equation}
\label{e:e4}
D_{ij} = F(L_{ij}) = \dfrac{c}  {\bar{L}} \times L_{ij},
\end{equation}
where $D_{ij}$ is the $(i,j)^{th}$ element of matrix $D \in {\mathbf{R}}^2$, which is the output of the Mean Adjustment layer, and $\bar{L}$ is the mean of $L$. Then $D_{ij} \in [0\ 1]$ and the mean value of $D$, denoted by $\overline{D}$, will be equal to $c$. Finally, the Bernoulli distribution $Ber()$ is applied to binarize the values of $D$; that is,

\begin{equation}
\label{e:e5}
B_{ij} = Ber(D_{ij}) =
\begin{cases}
1, & p = D_{ij}\\
0, & p = 1 - D_{ij}\\
\end{cases}.
\end{equation}
Notice that
\begin{equation}
\label{e:e5.5}
\overline{B} \approx \frac{1}{N^{2}} \sum_{i=1}^{N}\sum_{j=1}^{N} E(B_{ij}) = \frac{1}{N^{2}} \sum_{i=1}^{N}\sum_{j=1}^{N} D_{ij} = c,
\end{equation}
where $N^2$ is the total number of pixels of $L$ and $E(B_{ij})$ is the expected value of $B_{ij}$. Therefore, $B$ is binary matrix with mean value equal to $c$, implying that it has $c$ percent of $1$'s.

Since applying the function $Ber(F(\cdot))$ on the input $L$ will make the output of the network be binary and have $c$ percent of $1$'s, we then make it the last Binarization layer activation function. Notice that function $Ber(D)$ is not continuous and its derivatives do not exist, making the back propagation optimization during training impractical. We use its expected value $D$ to approximate it during training and apply the original function $Ber(D)$ during testing.

\section{Spatial-Spectral Representation for X-Ray Fluorescence Image Inpainting}
\label{sec:spatialspectralXRFinpainting}

In this section, we propose the XRF image inpainting algorithm by fusing it with a conventional RGB image, detailing the ``Proposed Inpainting Algorithm'' block in Figure~\ref{fig:XRFInpaintingDiagram}. The proposed fusion style inpainting approach has similarities with our previous fusion style SR approach~\cite{dai2017x}. We first formulate the XRF image inpainting problem, then demonstrate our proposed solution to this inpainting problem.

\subsection{Problem Formulation}
\label{sec:inpaintingproblemformulation}

As shown in Figure~\ref{fig:XRFInpaintingVNVPipeline}, we are seeking the estimation of a reconstructed XRF image $\bar{Y} \in \mathbb{R}^{W \times H \times B}$ that is fully sampled, with $W$, $H$, and $B$ the image width, height, and number of spectral bands, respectively. We have two inputs: a subsampled XRF image $\bar{X} \in \mathbb{R}^{W \times H \times B}$ with the known binary sampling mask $\bar{S} \in \mathbb{R}^{W \times H}$ ($\overline{X}(i,j,:)$ is equal to the zero vector if not sampled, i.e., corresponding to $\overline{S}(i,j) = 0$), and a conventional RGB image $\bar{I} \in \mathbb{R}^{W \times H \times b}$ with the same spatial resolution as the target XRF image $\bar{Y}$, but a small number (equal to $3$) of spectral bands, i.e., $b \ll B$. {Scanned data $\bar{X}$ and $\bar{I}$ is linearly scaled to $[0\ 1]$, i.e., $0 \leq \overline{X}(i,j,k) \leq 1$ and $0 \leq \overline{I}(i,j,k) \leq 1$.} Note that the RGB image is fully sampled, and therefore the primary goal of the reconstruction algorithm is to transfer image information from the RGB image to regions of the XRF where no samples are acquired. The input subsampled XRF image $\bar{X}$ can be separated into two parts: the visible component $\bar{X}_v \in \mathbb{R}^{W \times H \times B}$ and the non-visible component $\bar{X}_{nv} \in \mathbb{R}^{W \times H \times B}$, with the same binary sampling mask $\bar{S}$ as $\bar{X}$. We propose to estimate the fully sampled visible component $\bar{Y} \in \mathbb{R}^{W \times H \times B}$ by fusing the conventional RGB image $\bar{I}$ with the visible component of the input subsampled XRF image $\bar{X}_v$, and the fully sampled non-visible component $\bar{Y}_{nv} \in \mathbb{R}^{W \times H \times B}$ by using standard total variation inpainting methods.

To simplify notation, the image cubes are written as matrices, i.e., all pixels of an image are concatenated, such that every column of the matrix corresponds to the spectral response at a given pixel, and every row corresponds to a lexicographically ordered spectral band. Those unsampled pixels are skipped in this matrix representation. Accordingly, the image cubes are written as $Y \in \mathbb{R}^{B \times N_h}$, $X \in \mathbb{R}^{B \times N_s}$, $I \in \mathbb{R}^{b \times N_h}$, $X_v \in \mathbb{R}^{B \times N_s}$, $X_{nv} \in \mathbb{R}^{B \times N_s}$, $Y_v \in \mathbb{R}^{B \times N_h}$, $Y_{nv} \in \mathbb{R}^{B \times N_h}$, where $N_h = W \times H$ and $N_s = W \times H \times c$ is the number of sampled XRF pixels. We therefore have

\begin{equation}
\label{e:e6}
X = X_v + X_{nv},
\end{equation}

\begin{equation}
\label{e:e7}
Y = Y_v + Y_{nv},
\end{equation}
according to the visible/non-visible component separation models as shown in Figure~\ref{fig:XRFInpaintingVNVPipeline}.

{Let us denote by $y \in \mathbb{R}^B$, $y_v \in \mathbb{R}^B$, and $y_{nv} \in \mathbb{R}^B$ the one-dimensional spectra at the same pixel location of $\bar{Y}$, $\bar{Y}_{v}$ and $\bar{Y}_{nv}$, respectively.} That is, a column of $Y_v$ and $Y_{nv}$ is represented according to the linear mixing model~\cite{bioucas2012hyperspectral, keshava2002spectral} described as

\begin{equation}
\label{e:e8}
y_v = \sum_{j=1}^{M} d^{xrf}_{v,j} \alpha_{v,j},\ \ \  Y_v = D_v^{xrf}A_v,
\end{equation}

\begin{equation}
\label{e:e9}
y_{nv} = \sum_{j=1}^{M} d^{xrf}_{nv,j} \alpha_{nv,j},\ \ \  Y_{nv} = D_{nv}^{xrf}A_{nv},
\end{equation}
where $d^{xrf}_{v,j}$ and $d^{xrf}_{nv,j}$ are column vectors representing respectively the endmembers for the visible and non-visible components, $M$ is the total number of endmembers, $D_{v}^{xrf}\equiv [d^{xrf}_{v\ 1}, d^{xrf}_{v\ 2},  \ldots, d^{xrf}_{v\ M}] \in \mathbb{R}^{B \times M}$, $D_{nv}^{xrf}\equiv [d^{xrf}_{nv\ 1}, d^{xrf}_{nv\ 2},  \ldots, d^{xrf}_{nv\ M}] \in \mathbb{R}^{B \times M}$, and $\alpha_{v,j}$ and $\alpha_{nv,j}$ are the corresponding per-pixel abundances. {Equation~\ref{e:e7} can be written per column, by utilizing the same column in each of the three matrices involved, that is $y=y_v+y_{nv}$.} We take the corresponding $\alpha_{v,j, j=1,...,M}$ and stack them into an $M \times 1$ column vector. This vector then becomes the $k^{th}$ column of the matrix $A_v \in \mathbb{R}^{M \times N_h}$. In a similar manner, we construct matrix $A_{nv} \in \mathbb{R}^{M \times N_h}$. The endmembers $D^{xrf}_{v}$ and $D^{xrf}_{nv}$ act as basis dictionaries representing $Y_{v}$ and $Y_{nv}$ in a lower-dimensional space $\mathbb{R}^M$, with $rank\{Y_v\} \leq M, and\ rank\{Y_{nv}\} \leq M$.

The visible $X_v$ and non-visible $X_{nv}$ components of the input subsampled XRF image are spatially subsampled versions of $Y_{v}$ and $Y_{nv}$ respectively; that is

\begin{equation}
\label{e:e10}
X_{v}=Y_{v}S=D^{xrf}_{v}A_{v}S,
\end{equation}

\begin{equation}
\label{e:e11}
X_{nv}=Y_{nv}S=D^{xrf}_{nv}A_{nv}S,
\end{equation} where $S \in \mathbb{R}^{N_h \times N_s}$ is the subsampling operator that describes the spatial degradation from the fully sampled XRF image to the subsampled XRF image.

Similarly, the input RGB image $I$ can be described by the linear mixing model~\cite{bioucas2012hyperspectral, keshava2002spectral},

\begin{equation}
\label{e:e12}
I=D^{rgb} A_{v},
\end{equation} where $D^{rgb} \in \mathbb{R}^{b \times M}$ is the RGB dictionary. Notice that the same abundance matrix $A_v$ is used in Equations~\ref{e:e8} and~\ref{e:e10}. This is because the visible component of the scanning object is captured by both the XRF and the conventional RGB images. Matrix $A_v$ encompasses the spectral correlation between the visible component of the XRF and the RGB images.

\subsection{Proposed Solution}
\label{sec:proposedsolution}

To solve the XRF image inpainting problem, we need to estimate $A_v$, $A_{nv}$, $D^{rgb}$, $D^{xrf}_{v}$ and $D^{xrf}_{nv}$ simultaneously. Utilizing Equations~\ref{e:e6}, \ref{e:e10}, \ref{e:e11}, and \ref{e:e12}, we formulate the following constrained least-squares problem:

\begin{subequations}
\label{e:e14}
\begin{align}
\begin{split}
    \displaystyle \min_{\mathclap{\substack{A_{v}, A_{nv}, D^{rgb},\\ D^{xrf}_{v},D^{xrf}_{nv}}}} &\ \ \ \ \|X-D^{xrf}_{v}A_{v}S- D^{xrf}_{nv}A_{nv}S\|_F^2  \\
    &\displaystyle +\gamma \| \nabla_{I} (D^{xrf}_{v}A_{v}) \|_F^2 + \lambda \| \nabla (D^{xrf}_{nv}A_{nv}) \|_F^2 \\
    &\displaystyle +\|I-D^{rgb}A_{v} \|_F^2 \\
    \end{split}\\
    \textrm{s.t.}\ \ \ & 0 \leq D^{xrf}_{v\ ij} \leq 1, \forall i,j \\
    & 0 \leq D^{xrf}_{nv\ ij} \leq 1, \forall i,j \\
    & 0 \leq D^{rgb}_{ij} \leq 1, \forall i,j \\
    & A_{v\ ij} \geq 0, \forall i,j \\
    & A_{nv\ ij} \geq 0, \forall i,j \\
    & \mathbf{1^{T}}(A_v+A_{nv})=\mathbf{1^{T}}, \\
    & \| A_v + A_{nv} \|_0 \leq s,
\end{align}
\end{subequations}
with $\|\cdot\|_F$ denoting the Frobenius norm and $\| \cdot \|_0$ the $\ell_0$ norm, i.e., the number of nonzero elements of the given matrix. $D^{xrf}_{v,ij}$, $D^{xrf}_{nv,ij}$, $D^{rgb}_{ij}$, $A_{v,ij}$, and $A_{nv,ij}$ are the $(i,j)$ elements of matrices $D^{xrf}_{v}$, $D^{xrf}_{nv}$, $D^{rgb}$, $A_{v}$, and $A_{nv}$ respectively, and $\mathbf{1^{T}}$ denotes a row vector of $1$'s compatible with the dimensions of $A_v$ and $A_{nv}$. Equations~\ref{e:e14}b, \ref{e:e14}c, and \ref{e:e14}d enforce the nonnegative, bounded spectrum constraints on endmembers, Equations~\ref{e:e14}e and \ref{e:e14}f enforce the nonnegative constraints on abundances, and Equation~\ref{e:e14}g enforces the constraint that the visible component abundances and non-visible component abundances for every pixel sum up to one. These physically grounded constraints from~\cite{lanaras2015hyperspectral} are shown to be effective in our previous work~\cite{dai2017x}, by making full use of the fact that the XRF endmembers are XRF spectra of individual materials, and the abundances are proportions of those endmembers. 

The first term in Equation~\ref{e:e14}a represents a measure of the fidelity to the subsampled XRF data $X$, the second term is the total variation (TV) regularizer of the visible component, the third term is the TV regularizer of the non-visible component, and the last term is the fidelity to the observed RGB image $I$. The TV regularizer of the visible component $\nabla_{I} (D^{xrf}_{v}A_{v})$ is defined as

\begin{equation}
\label{e:e15}
\begin{array}{l}
\| \nabla_{I} (D^{xrf}_{v}A_{v}) \|_F^2  \\
\displaystyle = \sum_{i=1}^{H-1}\sum_{j=1}^{W-1} w_{i,j}^{down}\|D_{v}^{xrf}\bar{A}_{v}(i,j,:)-D_{v}^{xrf}\bar{A}_{v}(i+1,j,:)\|_2^2 \\
\ \ \ \ \ \ \ \ \ \ \ \ +w_{i,j}^{right}\|D_{v}^{xrf}\bar{A}_{v}(i,j,:)-D_{v}^{xrf}\bar{A}_{v}(i,j+1,:)\|_2^2 \\
\displaystyle = \| D_{v}^{xrf} A_{v} P(I)\|_F^2,
\end{array}
\end{equation}
where $\bar{A}_{v} \in \mathbb{R}^{W \times H \times M} $ is the $3D$ volume version of $A_{v}$ and $\bar{A}_{v}(i,j,:) \in \mathbb{R}^{M}$ is the {visible} component abundance of pixel $(i,j)$. $w_{i,j}^{down}$ and $w_{i,j}^{right}$ are the adaptive TV weights in the vertical and horizontal directions respectively; that is,

\begin{equation}
\label{e:e16}
w_{i,j}^{down}  = e^{-\alpha \| \bar{I}(i,j,:) - \bar{I}(i+1,j,:)\|_2^2},
\end{equation}

\begin{equation}
\label{e:e17}
w_{i,j}^{right} = e^{-\alpha \| \bar{I}(i,j,:) - \bar{I}(i,j+1,:)\|_2^2},
\end{equation}
where $\bar{I}(i,j,:)$ is the RBG image pixel at position $(i,j)$. $P(I) \in \mathbb{R}^{N_h \times ((W-1)(H-1))}$ in Equation~\ref{e:e15} is the adaptive horizontal/vertical first order difference operator determined by the input RGB image $I$ according to Equations~\ref{e:e16} and~\ref{e:e17}. Equations~\ref{e:e16} and~\ref{e:e17} indicate that the TV regularizer of the visible component adapts to the dense set of information that is available in the conventional RGB image $\bar{I}$. When the difference between two adjacent RGB pixels is small, a strong spatial smoothness constraint is placed on their corresponding XRF pixels, and vice versa. This adaptive TV regularizer is one of the main differences between this fusion-based XRF image inpainting algorithm and our previous fusion-based XRF image SR algorithm~\cite{dai2017x}. We found out that such TV regularizer on the visible component is essential for the inpainting problem, otherwise the inpainting results are not satisfactory. For the SR approach, we do not need such a TV regularizer on the visible component. The SR degradation model assumes that the LR measured XRF image is a weighted sum of all the pixels in the target HR XRF image, so there is an implicit spatial smoothness constraint imposed by the LR XRF image. However, for the XRF image inpainting problem, we subsample the XRF image to obtain the measurement so that many pixels are not sampled at all, making the reconstruction more difficult than for the SR problem. 

{The third term in Equation~\ref{e:e14} is a TV regularizer of the non-visible component. Its detailed definition can be found in our previous work~\cite{dai2017x} (see Equation (10)).}

The optimization in Equation~\ref{e:e14} is non-convex and difficult to carry out if we are to optimize over all the parameters $A_v$, $A_{nv}$, $D^{rgb}$, $D^{xrf}_{v}$, and $D^{xrf}_{nv}$ directly. {Empirically we found out that it is effective to alternatively optimize over these parameters.} Also, because Equation~\ref{e:e14} is highly non-convex, a good initialization is needed. Let $Y^{(0)} \in  \mathbb{R}^{B \times N_h}$ be the initialization of $Y$. Such initialization can be obtained by utilizing some standard image inpainting algorithms~\cite{esedoglu2002digital, zhou2012nonparametric} to inpaint the subsampled XRF image channel by channel. Then, the coupled dictionary learning technique in~\cite{yang2008image, yang2010image} can be utilized to initialize $D^{rgb}$ and $D^{xrf}_v$ by

\begin{equation}
\label{e:e19}
\begin{array}{ll}
\displaystyle\min_{\mathclap{\substack{ D^{rgb}, D_{v}^{xrf}}}} & \ \ \ \| I - D^{rgb}A_{v} \|_F^2 + \| Y^{(0)} - D^{xrf}_v A_{v} \|_F^2 \\
&\displaystyle \ \ \  + \beta \sum_{k=1}^{N_l}\| A_v(:,k) \|_1, \\
\textrm{s.t.} & \|D^{rgb}(:,k)\|_2 \leq 1, \forall k, \\
              & \|D^{xrf}_{v}(:,k)\|_2 \leq 1, \forall k,\\
\end{array}
\end{equation} where $\| \cdot \|_1$ is the $\ell_1$ vector norm, parameter $\beta$ controls the sparseness of the coefficients in $A_v$, and $A_{v}(:,k)$, $D^{rgb}(:,k)$, and $D^{xrf}_{v}(:,k)$ denote the $k^{th}$ column of matrices $A_v$, $D^{rgb}$, and $D^{xrf}_{v}$ respectively. Details of the optimization can be found in~\cite{yang2008image, yang2010image}. $D^{rgb}$ and $D_{v}^{xrf}$ are initialized using Equation~\ref{e:e19} and $D_{nv}^{xrf}$ is initialized to be equal to $D_{v}^{xrf}$. $A_v$ is initialized by Equation~\ref{e:e19} as well, while $A_{nv}$ is set equal to zero at initialization. Note that this formulation allows our approach to naturally handle the problem when no hidden layers are present. The iterative optimization algorithms we deploy are similar to the ones used in our previous work~\cite{dai2017x}, and they are not repeated here.

Once the optimization problem in Equation~\ref{e:e14} is solved according to Equations~\ref{e:e7}, \ref{e:e8}, and \ref{e:e9}, the reconstructed fully sampled XRF output image $Y$ can be computed by

\begin{equation}
\label{e:e28}
Y = Y_{v} + Y_{nv}=D_{v}^{xrf}A_v+D_{nv}^{xrf}A_{nv}.
\end{equation}

\begin{table*}[h]
\scriptsize
\begin{tabular}{c | c | c | c | c | c | c | c | c | c | c | c | c | c | c | c} \hline
\multirow{2}{*}{} & \multicolumn{5}{c|}{c=0.05} & \multicolumn{5}{c|}{c=0.1} & \multicolumn{5}{c}{c=0.2} \\ \hline
 & NetE & Harmonic & Mum-Sh  & BPFA & time(s) & NetE & Harmonic & Mum-Sh & BPFA & time(s) & NetE & Harmonic &  Mum-Sh & BPFA & time(s) \\ \hline
Random & 18.44 & 18.15 & 18.56 & 15.42 & - & 19.94 & 19.88 & 20.08 & 19.01 & - & 22.06 & 21.76 & 22.08 & 21.80 & - \\ \hline
AIrS & 17.27 & 16.72 & 18.01 & 13.24 & 0.22 & 18.50 & 18.52 & 19.53 & 16.56 & 0.23 & 18.22 & 20.30 & 20.60 & 18.49 & 0.24 \\ \hline
KbAS & 18.41 & 17.84 & 19.79 & 14.64 & 23.37 & 20.58 & 20.88 & 21.98 & 18.85 & 50.87 & 23.01 & 23.63 & 24.51 & 22.52 & 104.55 \\ \hline
Mascar & 18.39 & 17.44 & 19.20 & 14.67 & 0.04 & 20.18 & 19.95 & 20.87 & 19.30 & 0.04 & 21.63 & 22.27 & 23.13 & 21.53 & 0.04 \\ \hline
$NetM$ & \textbf{20.35} & \textbf{20.22} & \textbf{20.70} & \textbf{15.82} & \textbf{0.02} & \textbf{21.93} & \textbf{22.42} & \textbf{22.98} & \textbf{20.82} & \textbf{0.02} & \textbf{24.31} & \textbf{24.38} & \textbf{25.04} & \textbf{24.16} & \textbf{0.02} \\ \hline

\end{tabular}
\caption{Sampling and inpainting results on ImageNet test images. Random, AIrS~\cite{ramponi2001adaptive}, KbAS~\cite{liu2014kernel}, Mascar~\cite{taimori2018adaptive} and proposed $NetM$ sampling strategies are compared in PSNR under $NetE$~\cite{gao2016one}, Harmonic~\cite{chan2001nontexture}, Mum-Sh~\cite{esedoglu2002digital}, and BPFA~\cite{zhou2012nonparametric} inpainting algorithms. Computation time of different sampling strategies is also compared. Best results are shown in bold. The results shown are averaged over a set of 1000 test images.}
\label{table:rgb_sampling_inpainting_table}
\end{table*}

\section{Experimental Results}
\label{sec:experimenntalresults}

In this section, we will first demonstrate the advantages of our developed adaptive sampling mask when applied to RGB image sampling and reconstruction. We will then demonstrate the performance of the proposed fusion-based XRF inpainting algorithm.

\subsection{Adaptive Sampling Mask for RGB Image Sampling and Reconstruction}
\label{sec:rgbinpainting}

For the RGB image sampling and reconstruction task, we show the benefits of our developed adaptive sampling mask, over the use of a random sampling mask, as well as other adaptive sampling algorithms, such as Adaptive Irregular Sampling (AIrS) proposed in~\cite{ramponi2001adaptive}, Kernel-based Adaptive Sampling (KbAS) proposed in~\cite{liu2014kernel} {and Measurement-Adaptive Sampling and Cellular Automaton Recovery (Mascar) proposed in~\cite{taimori2018adaptive}}.

\subsubsection{Datasets}
\label{sec:expDatasets}

To train our proposed adaptive sampling mask generation CNN in Section~\ref{sec:cnnadapsampling},  the ImageNet~\cite{deng2009imagenet} database without any of the accompanying labels is used. We randomly select $1,000,000$ images for the training set, {$1000$} for the validation set, and {$1000$} for the testing set. All images are selected randomly among all categories to capture as diverse image structures as possible and are cropped to have spatial resolution $64 \times 64$ pixel.

\subsubsection{Implementation Details}
\label{sec:implementationDetails}

\begin{figure*}[h!]
\scriptsize
\begin{tabular}{ c c c c c c c c}

\shortstack{Test Image \\ \ } &  \shortstack{Sampling \\ Mask}  & \shortstack{Sampled\\Image} & \shortstack{NetE \\ Reconstruction } & \shortstack{Harmonic\\ Reconstruction} & \shortstack{Mum-Sh \\Reconstruction} & \shortstack{BPFA \\Reconstruction} \\ \hline \hline

\rotatebox{90}{\#39}
\includegraphics[width=0.1\textwidth]{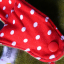}        &
\rotatebox{90}{Random}
\includegraphics[width=0.1\textwidth]{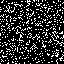}  &
\includegraphics[width=0.1\textwidth]{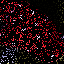}  &
\includegraphics[width=0.1\textwidth]{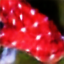}  &
\includegraphics[width=0.1\textwidth]{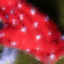}  &
\includegraphics[width=0.1\textwidth]{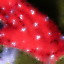}  &
\includegraphics[width=0.1\textwidth]{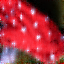}  \\

                                                                               &
                                                                               \rotatebox{90}{AIrS}
\includegraphics[width=0.1\textwidth]{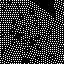}  &
\includegraphics[width=0.1\textwidth]{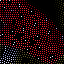}  &
\includegraphics[width=0.1\textwidth]{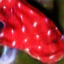}  &
\includegraphics[width=0.1\textwidth]{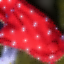}  &
\includegraphics[width=0.1\textwidth]{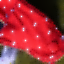}  &
\includegraphics[width=0.1\textwidth]{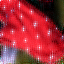}  \\

                                                                                &
                                                                                \rotatebox{90}{KbAS}
\includegraphics[width=0.1\textwidth]{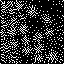}  &
\includegraphics[width=0.1\textwidth]{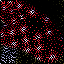}  &
\includegraphics[width=0.1\textwidth]{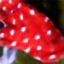}  &
\includegraphics[width=0.1\textwidth]{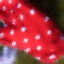}  &
\includegraphics[width=0.1\textwidth]{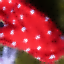}  &
\includegraphics[width=0.1\textwidth]{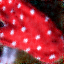}  \\

                                                                                &
                                                                                \rotatebox{90}{Mascar}
\includegraphics[width=0.1\textwidth]{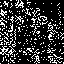}  &
\includegraphics[width=0.1\textwidth]{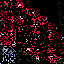}  &
\includegraphics[width=0.1\textwidth]{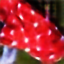}  &
\includegraphics[width=0.1\textwidth]{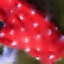}  &
\includegraphics[width=0.1\textwidth]{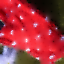}  &
\includegraphics[width=0.1\textwidth]{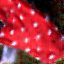}  \\

                                                                                &
                                                                                \rotatebox{90}{$NetM$}
\includegraphics[width=0.1\textwidth]{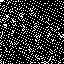}  &
\includegraphics[width=0.1\textwidth]{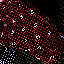}  &
\includegraphics[width=0.1\textwidth]{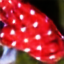}  &
\includegraphics[width=0.1\textwidth]{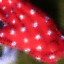}  &
\includegraphics[width=0.1\textwidth]{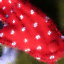}  &
\includegraphics[width=0.1\textwidth]{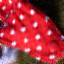}  \\

\rotatebox{90}{\#91}
\includegraphics[width=0.1\textwidth]{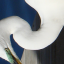}        &
\rotatebox{90}{Random}
\includegraphics[width=0.1\textwidth]{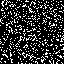}  &
\includegraphics[width=0.1\textwidth]{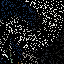}  &
\includegraphics[width=0.1\textwidth]{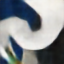}  &
\includegraphics[width=0.1\textwidth]{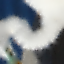}  &
\includegraphics[width=0.1\textwidth]{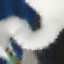}  &
\includegraphics[width=0.1\textwidth]{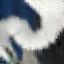}  \\

                                                                               &
                                                                               \rotatebox{90}{AIrS}
\includegraphics[width=0.1\textwidth]{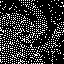}  &
\includegraphics[width=0.1\textwidth]{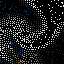}  &
\includegraphics[width=0.1\textwidth]{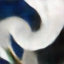}  &
\includegraphics[width=0.1\textwidth]{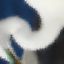}  &
\includegraphics[width=0.1\textwidth]{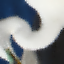}  &
\includegraphics[width=0.1\textwidth]{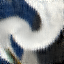}  \\
                                                                               &
                                                                               \rotatebox{90}{KbAS}
\includegraphics[width=0.1\textwidth]{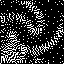}  &
\includegraphics[width=0.1\textwidth]{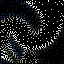}  &
\includegraphics[width=0.1\textwidth]{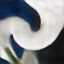}  &
\includegraphics[width=0.1\textwidth]{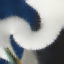}  &
\includegraphics[width=0.1\textwidth]{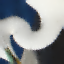}  &
\includegraphics[width=0.1\textwidth]{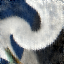}  \\
                                                                               &
                                                                               \rotatebox{90}{Mascar}
\includegraphics[width=0.1\textwidth]{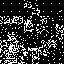}  &
\includegraphics[width=0.1\textwidth]{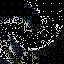}  &
\includegraphics[width=0.1\textwidth]{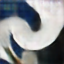}  &
\includegraphics[width=0.1\textwidth]{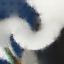}  &
\includegraphics[width=0.1\textwidth]{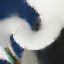}  &
\includegraphics[width=0.1\textwidth]{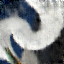}  \\
                                                                               &
                                                                               \rotatebox{90}{$NetM$}
\includegraphics[width=0.1\textwidth]{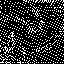}  &
\includegraphics[width=0.1\textwidth]{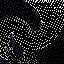}  &
\includegraphics[width=0.1\textwidth]{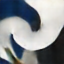}  &
\includegraphics[width=0.1\textwidth]{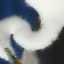}  &
\includegraphics[width=0.1\textwidth]{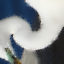}  &
\includegraphics[width=0.1\textwidth]{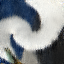}  \\

\end{tabular}

\caption{Visual Comparison of the reconstructed images using random, AIrS, KbAS, and $NetM$ sampling masks at sampling rate $c=0.2$. The first column is the input test image and the second column is the sampling mask, either random, AIrS, KbAS, or $NetM$, the third column is the sampled image obtained by the sampling mask, and the rest of the columns are the reconstruction results of $NetE$ Inpainting~\cite{gao2016one}, Harmonic Inpainting~\cite{chan2001nontexture}, Mumford-Shah Inpainting~\cite{esedoglu2002digital}, and BPFA inpainting~\cite{zhou2012nonparametric} respectively.}
\label{fig:imageNet_test_visual}

\end{figure*}

\begin{figure*}[h!]
\scriptsize
\begin{tabular}{c c c c c}

Test Image&  Sampling Mask  & Harmonic Reconstruction & Mum-Sh Reconstruction & BPFA Reconstruction \\ \hline \hline

\rotatebox{90}{Original RGB Image }
\includegraphics[width=0.15\textwidth]{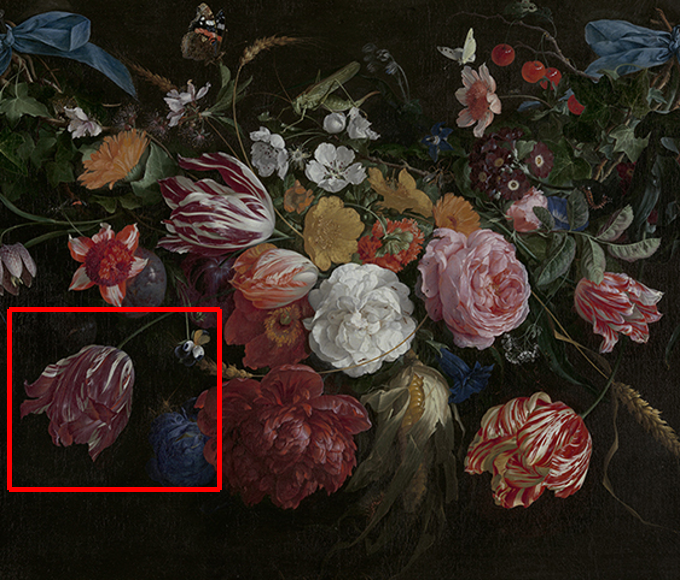}   &
\rotatebox{90}{Random}
\includegraphics[width=0.15\textwidth]{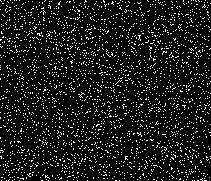}    &
\includegraphics[width=0.15\textwidth]{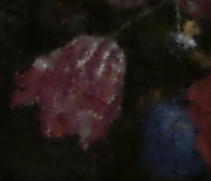}  &
\includegraphics[width=0.15\textwidth]{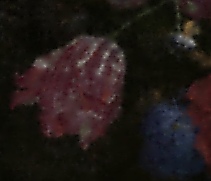} &    
\includegraphics[width=0.15\textwidth]{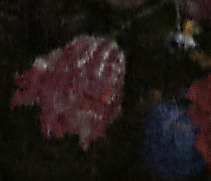}  \\                                                                     

(a) &
(c) time: - &
(d) PSNR: 25.85 dB &
(e) PSNR: 25.07 dB &
(f) PSNR: 24.59 dB \\

\rotatebox{90}{Cropped Original}
\includegraphics[width=0.15\textwidth]{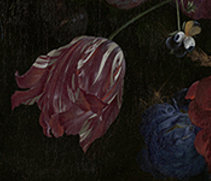}   &
\rotatebox{90}{AIrS}
\includegraphics[width=0.15\textwidth]{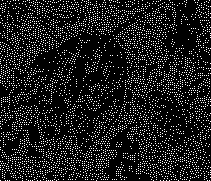}         & 
\includegraphics[width=0.15\textwidth]{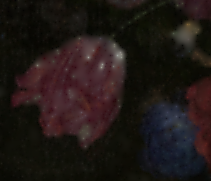}  &
\includegraphics[width=0.15\textwidth]{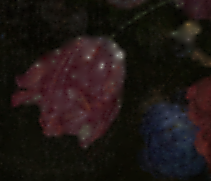}  &
\includegraphics[width=0.15\textwidth]{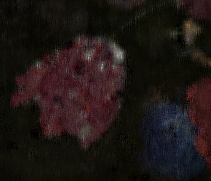}    \\

(b) &
(g) time: 468.15s&
(h) PSNR: 23.34 dB &
(i) PSNR: 23.24 dB &
(j) PSNR: 21.65 dB \\

&
\rotatebox{90}{KbAS}
\includegraphics[width=0.15\textwidth]{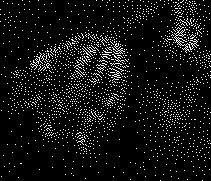}         & 
\includegraphics[width=0.15\textwidth]{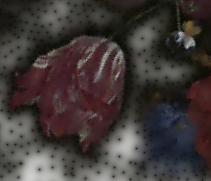}  &
\includegraphics[width=0.15\textwidth]{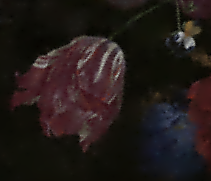}  &
\includegraphics[width=0.15\textwidth]{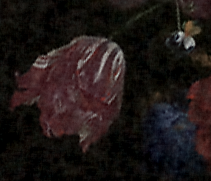}    \\

&
(k) time: 32533.10s&
(l) PSNR: 14.41 dB &
(m) PSNR: 28.43 dB &
(n) PSNR: 27.48 dB \\

&
\rotatebox{90}{Mascar}
\includegraphics[width=0.15\textwidth]{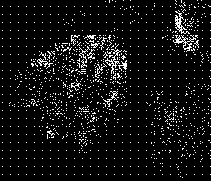}         & 
\includegraphics[width=0.15\textwidth]{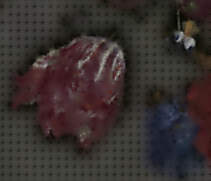}  &
\includegraphics[width=0.15\textwidth]{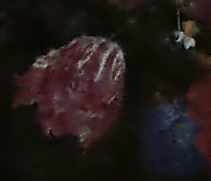}  &
\includegraphics[width=0.15\textwidth]{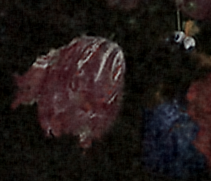}    \\

&
(o) time: 3.08s&
(p) PSNR: 19.77 dB &
(q) PSNR: 27.22 dB &
(r) PSNR: 26.83 dB \\

&
\rotatebox{90}{$NetM$}
\includegraphics[width=0.15\textwidth]{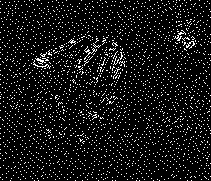}         & 
\includegraphics[width=0.15\textwidth]{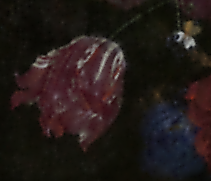}  &
\includegraphics[width=0.15\textwidth]{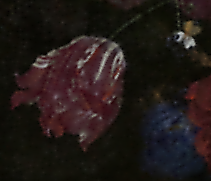}  &
\includegraphics[width=0.15\textwidth]{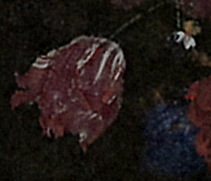}    \\

&
(s) time: 0.26s&
(t) PSNR: 28.13 dB &
(u) PSNR: 28.57 dB &
(v) PSNR: 28.57 dB \\

\end{tabular}
\caption{Visualization of sampling and inpainting result of the ``Bloemen en insecten'' painting. (a) original RGB image with red bounding box. (b) region inside the bounding box of (a) for visualization purposes. (c), (g), (k), (o) and (s) random, AIrS, KbAS, Mascar and $NetM$ sampling masks respectively. (d), (h), (l), (p) and (t) reconstruction results of each sampling mask using Harmonic algorithms. (e), (i), (m), (q) and (u) reconstruction results of each sampling mask using Mumford-Shah algorithm. (f), (j), (n), (r) and (v) reconstruction results of each sampling mask using BPFA algorithm. Computation time of each sampling mask and PSNR of the entirety of each reconstructed images are also shown.}
\label{fig:RGB_inpaint_visual_compare_DeHeem}
\end{figure*}

Our proposed adaptive sampling mask generation CNN (Section~\ref{sec:cnnadapsampling}) is implemented in PyTorch\footnote{\url{https://github.com/usstdqq/deep-adaptive-sampling-mask}}. ADAM~\cite{kingma2014adam} is applied as the stochastic gradient descent solver for optimization. We use the same hyper-parameters suggested in~\cite{gao2016one} and batch size equal to $128$ during training. $400$ epochs are applied during the training process. We test 3 sampling rates: $c = 5\%$,  $c = 10\%$, and  $c = 20\%$ for the sampling percentage parameter. A $5\%$, $10\%$, and $20\%$ sampling percentage roughly speed up the raster scanning procedure by a factor of $20$, $10$, and $5$, respectively.

For training, we first initialize the inpainting network $NetE$ according to~\cite{gao2016one}. A random sampling mask with sampling rate $c$ is utilized to sample the input RGB image. The mask generation network $NetM$ is initialized randomly. We then train the whole network architecture in Figure~\ref{fig:pipelineadaptive}. The learning rate of the mask generation network $NetM$ is set equal to $0.0002$ during training. The learning rate of the inpainting network $NetE$ is set to $0$, i.e., we fix $NetE$ when training $NetM$. We did not optimize $NetM$ and $NetE$ simultaneously; although the best reconstruction of $z$ would have been obtained, the two networks would have been dependent on each other. Notice that the channel-wise fully connected layer in $NetE$ (Figure~\ref{fig:inpaintingNet}) is able to learn a high-level feature mapping, making $NetE$ able to perform semantic image inpainting. However, we would like to utilize other image inpainting algorithms other than $NetE$ to make the adaptive sampling mask be as general and applicable to as many image inpainting algorithms as possible. {Also as mentioned in Section~\ref{sec:maskarchitecture}, during training, the output of $NetM$ is $D \in {\mathbf{R}}^2$, a probabilistic map, instead of the binary mask $Ber(D)$. So training $NetE$ with $NetM$ jointly would make $NetE$ learn to reconstruct images corrupted by multiplying them by a continuous probabilistic mask, instead of a binary mask, which does not represent the inpainting problem. By fixing $NetE$, which is pre-trained by random binary sampling masks, $NetM$ is constrained to be optimized for the general image inpainting problem instead of the ``probabilistic'' image inpainting problem.} An adaptive sampling mask is thus trained suitable for general image inpainting problem. Note that, since $NetE$ architecture is differentiable, it allows us to train the $NetM$ atchitecture to optimize the mask, while taking into account both the RGB image content, and the reconstruction algorithm.

%\subsubsection{Baselines}
%\label{sec:expBaselines}

\subsubsection{Performance on ImageNet Testing Images}
\label{sec:expRGBInpaintingImageNet}

To compare the performance of our adaptive sampling mask ($NetM$) with the random, AIrs~\cite{ramponi2001adaptive}, KbAS~\cite{liu2014kernel} {and Mascar~\cite{taimori2018adaptive} sampling masks, we apply all of them to corrupt  $1000$ testing images from the ImageNet database.} Three sampling rates  $c = 0.05$, $c = 0.1$ and $c = 0.2$ are tested for the sampling methods. For image inpainting algorithms, $NetE$ Inpainting~\cite{gao2016one}, Harmonic Inpainting~\cite{chan2001nontexture}, Mumford-Shah (Mum-Sh) Inpainting~\cite{esedoglu2002digital}, and BPFA inpainting~\cite{zhou2012nonparametric} are used to reconstruct the fully sampled RGB images.

%\begin{figure}[h]
%\centering
%\includegraphics[width=0.9\linewidth]{figs/result_ssim.pdf}
%\caption{Average SSIM over all 100 test images from the ImageNet database for several inpainting methods for random and adaptive sampling masks. }
%\label{fig:imageNet_test_ssim}
%\end{figure}

The average PSNRs in dB over all {$1000$} test images are shown in Table~\ref{table:rgb_sampling_inpainting_table}. {First, we observe that under all three sampling rates, the proposed $NetM$ mask outperforms the random, AIrS, KbAS and Mascar sampling masks consistently over all inpainting reconstruction algorithms in terms of PSNR, showing the effectiveness of our proposed adaptive sampling mask generation network. Furthermore, the proposed $NetM$ is significantly faster than AIrS, KbAS, and Mascar in generating the sampling mask.} Finally, it can be concluded that the smaller the sampling rate, the larger the advantage of our proposed algorithm compared to other sampling algorithms. This implies that our proposed $NetM$ is able to handle challenging sampling tasks (small sampling rates) and have better reconstruction accuracy under various reconstruction algorithms.

The visual quality comparison of the adaptive sampling and random sampling masks is shown in Figure~\ref{fig:imageNet_test_visual}. Two test images are picked from the testing set of {$1000$} images in total. {Under sampling rate $c=0.2$, the advantages of the proposed $NetM$ mask over the random, AIrS, KbAS and Mascar sampling masks can be observed by comparing the resulting reconstructions by the same inpainting algorithm.} For the test image $\#39$, the $NetM$ mask is able to capture the white dots in the red hat, resulting in accurate reconstruction results of those white dots. KbAS misses one white dot in the image. {Although AIrS and Mascar are able to sample all the white dots in the image, they fail to capture the structure of these white dots.} For test image $\#91$, compared to random sampling mask, the proposed $NetM$ mask samples the contour structure of the bird, resulting in its better reconstruction. {When compared to AIrS, KbAS and Mascar masks, the proposed $NetM$ mask samples the whole image more evenly, resulting in fewer artifacts in the inpainted images.} The improved performance of the adaptive sampling mask over other sampling masks is consistent over all inpainting reconstruction algorithms we tested.

\subsubsection{Performance on Painting Images}
\label{sec:expRGBInpaintingPainting}

We also tested our proposed adaptive sampling algorithm on painting images at sampling rate $c=0.1$. As shown in Figures~\ref{fig:RGB_inpaint_visual_compare_DeHeem} (a), the RGB image of the painting ``Bloemen en insecten'' is tested. It has spatial resolution $580 \times 680$ pixels. Random, AIrS, KbAS, Mascar and $NetM$ sampling masks are generated as shown in Figures~\ref{fig:RGB_inpaint_visual_compare_DeHeem}(c), (g), (k), (o) and (s) with the corresponding computation time. Harmonic Inpainting~\cite{chan2001nontexture}, Mumford-Shah Inpainting~\cite{esedoglu2002digital}, and BPFA~\cite{zhou2012nonparametric} algorithms are utilized to reconstruct the sampled RGB images, and the reconstruction results are shown in Figures~\ref{fig:RGB_inpaint_visual_compare_DeHeem} (d)-(f), (h)-(j), (l)-(n), (p)-(r) and (t)-(v) with the corresponding PSNR values. By comparing the rows of different sampling masks, it can be concluded that our proposed $NetM$ mask outperforms other sampling masks in terms of both visual quality of the reconstructed images and the PSNR values. Notice that $NetM$ is significantly faster than AIrS, KbAS and Mascar in computation speed. Although KbAS samples densely on the foreground, it still misses many details due to the complexity of the flower structure. {Similarly, Mascar samples densely on the foreground, while it misses the flower stem structure.} $NetE$ Inpainting~\cite{gao2016one} is not utilized in this experiment since it is trained to inpaint RGB images with spatial resolution $64 \times 64$ pixels. The network structure shown in Figure~\ref{fig:inpaintingNet} is not fully convolutional, as there is the channel-wise fully connected layer in the middle. {$NetM$ is fully convolutional (Figure~\ref{fig:maskNet}) so that it can generate sampling masks of input images with arbitrary resolution.}

\subsection{Adaptive Sampling Mask for X-Ray Fluorescence Image Inpainting}
\label{sec:xrfinpainting}

In the previous section (Section~\ref{sec:rgbinpainting}), we demonstrated the effectiveness of our proposed $NetM$ sampling mask on the RGB image sampling and inpainting problem. To further evaluate the effectiveness of the $NetM$ sampling mask and evaluate the performance of our proposed fusion-based inpainting algorithm (Section~\ref{sec:spatialspectralXRFinpainting}), we have performed experiments on XRF images. The basic parameters of the proposed reconstruction method are set as follows: the number of atoms in the dictionaries $D^{rgb}$, $D_{nv}^{xrf}$ and $D_{v}^{xrf}$ is $M=200$; parameter $\lambda$ and $\gamma$ in Equation~\ref{e:e14} are set equal to $0.1$; parameter $\alpha$ in Equation~\ref{e:e16} and Equation~\ref{e:e17} is set to $16$. The optional constraint in Equation~\ref{e:e14}h is not applied here.

\subsubsection{Error Metrics}
\label{sec:expXRFErrorMetric}

The root mean squared error (RMSE), the peak-signal-to-noise ratio (PSNR), and the  spectral angle mapper (SAM,~\cite{yuhas1992discrimination}) between the estimated fully sampled XRF image $Y$ and the ground truth image $Y^{gt}$ are used as the error metrics.

\subsubsection{Comparison Methods}
\label{sec:expXRFCompareMethod}

According to our knowledge, no work has been reported on solving the XRF (or Hyperspectral) image inpainting problem by fusing a conventional RGB image. So, we can only compare our results with traditional image inpainting algorithms such as  Harmonic Inpainting~\cite{chan2001nontexture}, Mumford-Shah Inpainting~\cite{esedoglu2002digital}, and BPFA inpainting~\cite{zhou2012nonparametric}. Harmonic Inpainting and Mumford-Shah Inpainting methods are for image inpainting, so we have to inpaint the XRF image channel by channel. BPFA inpainting~\cite{zhou2012nonparametric} is able to inpaint multiple channels simultaneously. {For the sampling mask comparison, we still compare with random, AIrS~\cite{ramponi2001adaptive}, KbAS~\cite{liu2014kernel} and Mascar~\cite{taimori2018adaptive} sampling masks.}

\subsubsection{Real Experiment}
\label{sec:expXRFReal}

For this experiment, real data were collected by a home-built X-ray fluorescence spectrometer (courtesy of Prof. Koen Janssens), with 2048 channels in spectrum. Studies from the XRF image scanned from Jan Davidsz. de Heem's ``Bloemen en insecten'' (ca 1645), in the collection of Koninklijk Museum voor Schone Kunsten (KMKSA) Antwerp, are presented here. We utilize the super-resolved XRF image in our previous work~\cite{dai2017x} as the ground truth. The ground truth XRF image has dimensions $680 \times 580 \times 2048$. We first extract 20 regions of interest (ROI) spectrally and work on them, to decrease the spectral dimension from $2048$ to $20$. We decrease the spectral dimension so as to compare with other inpainting algorithms, e.g.,~\cite{chan2001nontexture,esedoglu2002digital}, which therefore reconstruct the subsampled XRF image channel by channel and large spectral dimensions will make the computational time very long. The sampling ratio $c$ is set to be $0.05$, $0.1$, and $0.2$. Different sampling strategies are applied and analyzed. Various inpainting methods are subsequently applied to reconstruct those subsampled XRF images.

\begin{table*}[h]
\footnotesize
\begin{tabular}{c | c | c | c | c | c | c | c | c | c | c | c | c }\hline\hline
& \multicolumn{12}{c}{c=0.05} \\ \hline
& \multicolumn{3}{c|}{Harmonic} &  \multicolumn{3}{c|}{Mum-Sh} & \multicolumn{3}{c|}{BPFA} & \multicolumn{3}{c}{Proposed}\\ \hline
& RMSE & PSNR & SAM & RMSE & PSNR & SAM & RMSE & PSNR & SAM & RMSE & PSNR & SAM  \\ \hline
Random & 0.0428 & 27.37 & 5.06 & 0.0319 & 29.93 & 3.85 & 0.1141 & 22.57 & 8.27 & 0.0300 & 30.45 & 3.67 \\  \hline
AIrS & 0.1224 & 18.24 & 8.20 & 0.0378 & 28.45 & 4.33 & 0.1446 & 17.80 & 13.49 & 0.0355 & 37.03 & 4.14 \\ \hline
KbAS & 0.2374	& 12.49 &	12.62 & 0.0309 &	30.19 & 3.95 & 	0.1791 & 15.08 & 19.51 & 0.0297 & 39.25 & 3.85\\ \hline
Mascar & 0.2374	& 12.49 &	12.62 & 0.0309 &	30.19 & 3.95 & 	0.1791 & 15.08 & 19.51 & 0.0297 & 39.25 & 3.85\\ \hline
$NetM$ & \textit{0.0333} & \textit{29.56} & \textit{4.22} & \textit{0.0279} & \textit{31.09} & \textit{3.43} & \textit{0.0910} & \textit{24.22} & \textit{7.21} & \textit{\textbf{0.0261}} & \textit{\textbf{39.54}} & \textit{\textbf{3.27}} \\ \hline \hline
& \multicolumn{12}{c}{c=0.1} \\ \hline
& \multicolumn{3}{c|}{Harmonic} &  \multicolumn{3}{c|}{Mum-Sh} & \multicolumn{3}{c|}{BPFA} & \multicolumn{3}{c}{Proposed}\\ \hline
& RMSE & PSNR & SAM & RMSE & PSNR & SAM & RMSE & PSNR & SAM & RMSE & PSNR & SAM  \\ \hline
Random & 0.0260 & 31.70 & 3.10 & 0.0247 & 32.14 & 2.84 & 0.0369 & 28.71 & \textit{3.27} & 0.0229 & 32.80 & 2.66 \\  \hline
AIrS & 0.0408 & 27.78 & 3.77 & 0.0284 & 30.92 & 3.04 & 0.0703 & 23.12 & 4.30 & 0.0260 & 38.54 & 2.84 \\ \hline
KbAS & 0.1473 & 16.63 & 7.78 & 0.0231 & 32.74 & 2.83 & 0.1268 & 18.06 & 9.00 & 0.0220 & 41.59 & 2.73\\ \hline
Mascar & 0.2374	& 12.49 &	12.62 & 0.0309 &	30.19 & 3.95 & 	0.1791 & 15.08 & 19.51 & 0.0297 & 39.25 & 3.85\\ \hline
$NetM$ & \textit{0.0227} & \textit{32.87} & \textit{2.84} & \textit{0.0211} & \textit{33.53} & \textit{2.50} & \textit{0.0353} & \textit{29.66} & 3.61 & \textit{\textbf{0.0195}} & \textit{\textbf{41.87}} & \textit{\textbf{2.38}} \\ \hline \hline
& \multicolumn{12}{c}{c=0.2} \\ \hline
& \multicolumn{3}{c|}{Harmonic} &  \multicolumn{3}{c|}{Mum-Sh} & \multicolumn{3}{c|}{BPFA} & \multicolumn{3}{c}{Proposed}\\ \hline
& RMSE & PSNR & SAM & RMSE & PSNR & SAM & RMSE & PSNR & SAM & RMSE & PSNR & SAM  \\ \hline
Random & 0.0195 & 34.19 & 2.18 & 0.0184 & 34.70 & 1.92 & 0.0176 & 35.29 &	2.01 & 0.0168 & 35.48 & 1.79 \\  \hline
AIrS & 0.0185 & 34.66 & 1.94 & 0.0154 & 36.24 & 1.59 & 0.0336 & 29.50 & 2.48 & 0.0141 & 42.95 & 1.49 \\ \hline
KbAS & 0.0518 & 25.71 & 3.35 & 0.0157 & 36.08 & 1.78 & 0.0683 & 23.39 & 3.21 & 0.0148 & 44.02 & 1.72\\ \hline
Mascar & 0.2374	& 12.49 &	12.62 & 0.0309 &	30.19 & 3.95 & 	0.1791 & 15.08 & 19.51 & 0.0297 & 39.25 & 3.85\\ \hline
$NetM$ & \textit{0.0160} & \textit{35.90} & \textit{1.86} & \textit{0.0145} & \textit{36.77} & \textit{1.54} & \textit{0.0151} & \textit{36.70} &	\textit{1.80} & \textit{\textbf{0.0137}} & \textit{\textbf{44.10}} & \textit{\textbf{1.48}} \\ \hline

\end{tabular}
\caption{Experimental results on the ``Bloemen en insecten'' data comparing different inpainting methods, under random, AIrS, KbAS, and $NetM$ sampling strategies, discussed in Section~\ref{sec:expXRFCompareMethod} in terms of RMSE, PSNR, and SAM. Best sampling masks under the each reconstruction algorithm are shown in italic. Best reconstruction results are shown in bold.}
\label{table:deheem_compare_table}
\end{table*}

\begin{figure}[h!]
\centering
\includegraphics[angle=90, width=1.0\linewidth]{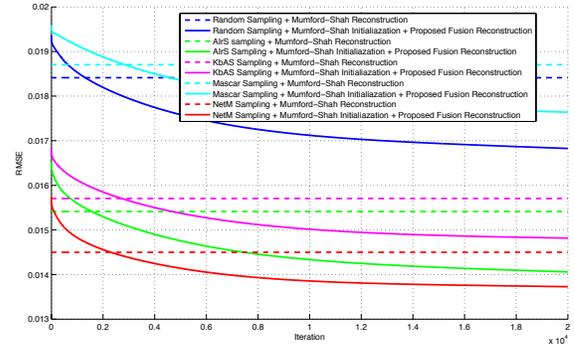}
\caption{RMSE versus number of iterations for our proposed fusion inpainting algorithm on the ``Bloemen en insecten'' data when $c=0.2$. The Mumford-Shah inpainting algorithm is utilized as initialization of our proposed algorithm. Different sampling masks are compared.}
\label{fig:Deheem_iter_process}
\end{figure}

\begin{figure*}[h!]
\scriptsize
\begin{tabular}{c c c c c}

Ground Truth & Harmonic  & Mum-Sh  & BPFA  & Proposed  \\ \hline\hline

\includegraphics[width=0.16\textwidth]{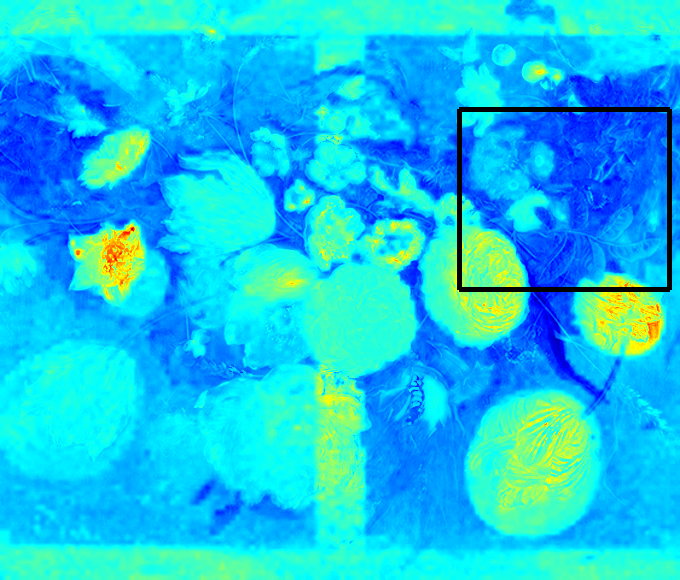}                                 &
\rotatebox{90}{Random Mask}
\includegraphics[width=0.16\textwidth]{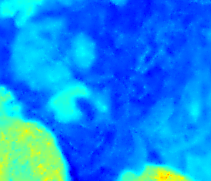}      &
\includegraphics[width=0.16\textwidth]{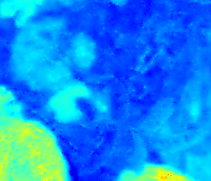}                             &
\includegraphics[width=0.16\textwidth]{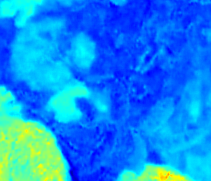}              &
\includegraphics[width=0.16\textwidth]{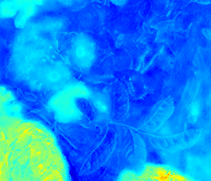}                        \\

(a) &
(c) &
(d) &
(e) &
(f) \\

\includegraphics[width=0.16\textwidth]{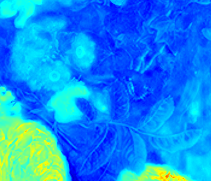} 	&
\rotatebox{90}{AIrS Mask}
\includegraphics[width=0.16\textwidth]{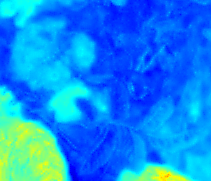}      &
\includegraphics[width=0.16\textwidth]{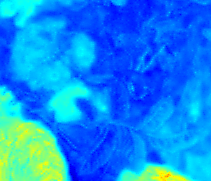}                             &
\includegraphics[width=0.16\textwidth]{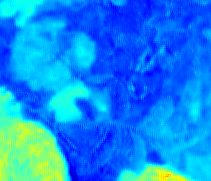}              &
\includegraphics[width=0.16\textwidth]{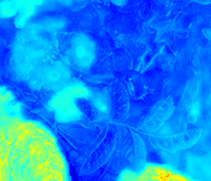}                        \\
    
(b) &
(g) &
(h) &
(i) &
(j) \\

&
\rotatebox{90}{KbAS Mask}
\includegraphics[width=0.16\textwidth]{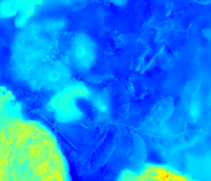}      &
\includegraphics[width=0.16\textwidth]{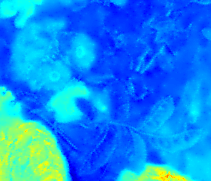}                             &
\includegraphics[width=0.16\textwidth]{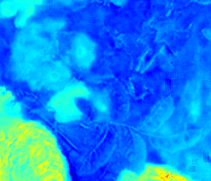}              &
\includegraphics[width=0.16\textwidth]{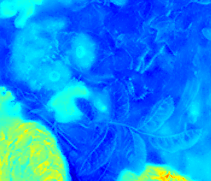}                         \\

&
(k) &
(l) &
(m) &
(n) \\

&
\rotatebox{90}{Mascar Mask}
\includegraphics[width=0.16\textwidth]{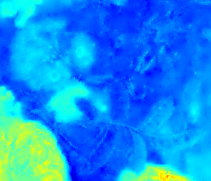}      &
\includegraphics[width=0.16\textwidth]{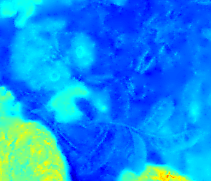}                             &
\includegraphics[width=0.16\textwidth]{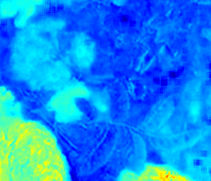}              &
\includegraphics[width=0.16\textwidth]{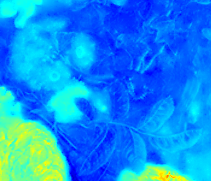}                         \\

&
(o) &
(q) &
(p) &
(r) \\

&
\rotatebox{90}{$NetM$ Mask}
\includegraphics[width=0.16\textwidth]{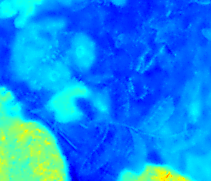}      &
\includegraphics[width=0.16\textwidth]{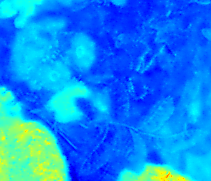}                             &
\includegraphics[width=0.16\textwidth]{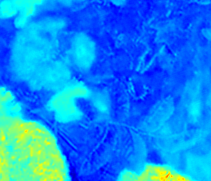}              &
\includegraphics[width=0.16\textwidth]{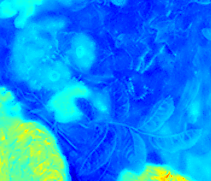}             \\

&
(s) &
(t) &
(u) &
(v) \\

\end{tabular}
\caption{Visualization of inpainting results on the ``Bloemen en insecten'' data when $c=0.2$. Channel \#16 related to the $Pb\ L\eta$ XRF emission line is selected. (a) ground truth XRF image with black bounding box. (b) region inside the bounding box of (a), shown for visual comparison purposes. The sampling masks for random, AIrS, KbAS, Mascar and $NetM$ are the same as the sampling masks in Figures~\ref{fig:RGB_inpaint_visual_compare_DeHeem} (c), (g), (k), (o) and (s) respectively. (c)-(v) reconstruction results of different inpainting algorithms for different sampled XRF image within the same region of (a) as (b).}
\label{fig:XRF_inpaint_visual_compare_Deheem}
\end{figure*}

As shown in Table~\ref{table:deheem_compare_table}, our proposed fusion-based inpainting algorithm with the proposed adaptive sampling mask provides the closest reconstruction to the ground truth XRF image compared to all other methods. Our proposed algorithm utilizes as guidance a conventional fully sampled and high contrast RGB image (Figure~\ref{fig:RGB_inpaint_visual_compare_DeHeem} (a)), resulting in better inpainting performance. By comparing the difference between results by ``Mum-Sh'' and results by ``Proposed'' under various sampling rates, it can be concluded that the benefit gained by our proposed fusion-based inpainting is large when the adaptive sampling masks are applied. {For example, at sampling rate $c=0.2$, there is a $0.78~dB$ improvement in PSNR by applying our proposed fusion-based inpainting algorithm when a random sampling masks are applied, while there is a  $6.71~dB$, $7.94~dB$, $9.06~dB$ and $7.33~dB$ improvement in PSNR when an AIrS, KbAS, Mascar and $NetM$ sampling mask is applied, respectively.} This is because the adaptive sampling masks sampled the corresponding visible component of the XRF image efficiently and the fusion inpainting propagated the measured XRF pixels properly. Furthermore, the $NetM$ sampling mask provides the best XRF image reconstruction compared to other sampling masks.

The iteration process of our proposed fusion inpainting algorithm when $c=0.2$ is shown in Figure~\ref{fig:Deheem_iter_process}. Notice that at the initial iterations of our proposed fusion inpainting algorithm, the RMSE is higher than the Mumford-Shah inpainting algorithm. This is because we decompose the inpainting result of Mumford-Shah inpainting algorithm by sparse representation, according to Equation~\ref{e:e19}. Due to the complexity of the ``Bloemen en insecten'' data, we lose some accuracy during the first few iterations. However, with more iterations, the RMSE of both random sampling and adaptive sampling decreases and becomes smaller than the RMSE of Mumford-Shah inpainting algorithm. Our proposed $NetM$ sampling mask provides the smallest RMSE for both Mumford-Shah initialization and the proposed fusion inpainting algorithm, showing the effectiveness of $NetM$ in the task of XRF image sampling and reconstruction. {In addition, the optimization of Equation~\ref{e:e14} exhibits monotonic convergence under all sampling masks.}

When $c=0.2$, the visual quality of the different inpainting algorithms and sampling strategies on channel \#16, corresponding to the $Pb\ L\eta$ XRF emission line, is compared in Figure~\ref{fig:XRF_inpaint_visual_compare_Deheem}. {The same random, AIrS, KbAS, Mascar and $NetM$ sampling masks as the sampling masks in Figures~\ref{fig:RGB_inpaint_visual_compare_DeHeem} (b), (f), (j), and (n), are applied here. The reconstruction results obtained by the Harmonic, Mum-Sh, BPFA, and Proposed algorithms using AIrS, KbAS, Mascar and $NetM$ sampling masks are sharper than those using the random sampling mask.} This is because the majority of the XRF signal in the ``Bloemen en insecten'' data correlates to the RGB signal, and the sampling masks, which are designed for the RGB image, would also be suitable for the visible component of the XRF signal. {Among the AIrS, KbAS, Mascar and $NetM$ sampling masks, $NetM$ sampling mask outperforms others for different reconstruction algorithms, both quantitatively (Table~\ref{table:deheem_compare_table}) and qualitatively (Figure~\ref{fig:XRF_inpaint_visual_compare_Deheem}).} The proposed fusion inpainting algorithm further improves the contrast and resolves more fine details in (q). When compared to the ground truth image (b), we can conclude that those resolved details have high fidelity to the ground truth image (b). 

Our proposed adaptive sampling algorithm $NetM$ is not designed and trained for the task of XRF image sampling and reconstruction. {However, $NetM$ still outperforms random sampling, as well as three other adaptive sampling algorithms, under various XRF reconstruction algorithms and under different sampling rates.} This illustrates the effectiveness of $NetM$ in extracting image information with limited sampling budget. This also demonstrates that $NetM$ can generalize well into some other imaging tasks.

{
\subsubsection{XRF Inpainting v.s. XRF Super-Resolution}
\label{sec:expInpaintingvsSR}

In this experiment, we further compare the proposed XRF inpainting utilizing $NetM$ adaptive sampling with our previous XRF-SR approach~\cite{dai2017x}. The XRF scanner can perform regular sub-sampling with a sampling step size $K$ on both image dimensions. An upscale factor $K$ SR can be applied to reconstruct the fully sampled XRF images. We set $K=5$ to match the SR setting of\cite{dai2017x}. We retrain $NetM$ with sampling rate $c=0.04$, which has the same number of samples as a $K=5$ regular sub-sampling. The training procedure is the same as described in Section~\ref{sec:implementationDetails}.

\begin{table}[h]
\footnotesize
\begin{tabular}{c | c | c | c }\hline
          & RMSE   & PSNR  & SAM \\ \hline
XRF-SR    & 0.0382 & 28.36 & 4.77 \\ \hline
Proposed  & \textbf{0.0278} & \textbf{31.12} & \textbf{3.55} \\ \hline
\end{tabular}
\caption{Experimental results on the ``Bloemen en insecten'' data comparing proposed XRF inpainting utilizing $NetM$ adaptive sampling with XRF SR. Best reconstruction results are shown in bold.}
\label{table:xrf_inpaintingvssr}
\end{table}

\begin{table}[h]
\footnotesize
\begin{tabular}{c | c | c | c }\hline
                                              & $NetE$   & Harmonic  & Mum-Sh \\ \hline
Random                                & 19.94      & 19.88         & 20.08      \\ \hline
AIrS                                       & 18.50      & 18.52         & 19.53       \\ \hline
KbAS                                     & 20.58     & 20.88         & 21.98       \\ \hline
Mascar                                  & 20.18      & 19.95         & 20.87       \\ \hline
$NetM\_pre\_train\_joint$    & 19.53      & 19.72         & 19.89        \\ \hline
$NetM\_rand\_init\_joint$    & 15.37      & 19.87         & 20.08        \\ \hline
$NetM\_sequential$            & \textbf{21.93}      & \textbf{22.42}        & \textbf{22.98}      \\ \hline
\end{tabular}
\caption{Sampling and inpainting results on ImageNet test images. Random, AIrS~\cite{ramponi2001adaptive}, KbAS~\cite{liu2014kernel}, Mascar~\cite{taimori2018adaptive},  $NetM\_pre\_train\_joint$ , $NetM\_rand\_init\_joint$ and proposed $NetM$ sampling strategies are compared in terms of PSNR (in dB) under $NetE$~\cite{gao2016one}, Harmonic~\cite{chan2001nontexture} and Mum-Sh~\cite{esedoglu2002digital} inpainting algorithms. Best results are shown in bold. The results shown are averaged over a set of 1000 test images.}
\label{table:simul_nete_netm}
\end{table}

As shown in Table~\ref{table:xrf_inpaintingvssr}, as expected, the proposed XRF inpainting algorithm utilizing $NetM$ outperforms the XRF-SR approach. Please notice that both methods fuse an RGB image during the XRF reconstruction, and the main difference is the sampling pattern. This experiment further argues the improved effectiveness of the proposed $NetM$ based sampling.

\subsubsection{Simultaneous Training of $NetE$ and $NetM$}
\label{sec:simultaneousNetENetM}

With an initial look at Fig.~\ref{fig:pipelineadaptive}, it would seem meaningful to train $NetE$ and $NetM$ simultaneously and not sequentially, as described in Section~\ref{sec:implementationDetails}. In this experiment, we performed certain experiments to test such approach and eventually support our proposed sequential training procedure. There are two approaches in training $NetE$ and $NetM$ simultaneously. The first one is to train $NetE$ individually first, and then train $NetE$ with $NetM$ jointly using Equation~\ref{e:e3}. The second one is to perform random initialization on both $NetE$ and $NetM$, and then simultaneously train them using Equation~\ref{e:e3}. We tested both of these training approaches at sampling rate $c=0.1$. We henceforth refer to the first approach as $NetM\_pre\_train\_joint$, to the second approach as $NetM\_rand\_init\_joint$ and to the proposed training approach as $NetM\_sequential$.  Notice that we quantize the probabilistic map $D$ generated by $NetM$ during testing.

We tested these two above described networks utilizing the same $1000$ ImageNet test set and compared them with the other inpainting approaches considered in this paper in terms of PSNR (in dB). As is clear from Table~\ref{table:simul_nete_netm}, the results obtained by the networks $NetM\_pre\_train\_joint$  and $NetM\_rand\_init\_joint$ , with which $NetM$ and $NetE$ were trained simultaneously, are not competitive with the results obtained by the proposed sequential training of $NetE$ and $NetM$, i.e., $NetM\_sequential$. It is mention again here that $NetE$ in both of these simultaneous training approaches was trained with input images corrupted by a probabilistic map $D$. The resulting sampling masks from these two approaches cannot also get good reconstruction accuracy using Harmonic~\cite{chan2001nontexture} and Mum-Sh~\cite{esedoglu2002digital} inpainting. There is approximately a $2$ dB difference in reconstruction accuracy, indicating that the trained $NetM$ is not as effective. When comparing these two simultaneous optimization approaches to each other, it is observed that using the pre-trained $NetE$ before joint training provides some benefit on both $NetE$ and $NetM$.

}

\section{Conclusion}
\label{sec:conclusion}

In this paper, we presented a novel adaptive sampling mask generation algorithm based on CNNs and a novel XRF image inpainting framework based on fusing a conventional RGB image. For the adaptive sampling mask generation, we trained the mask generation network $NetM$ along with the inpainting network $NetE$ to obtain an optimal binary sampling mask based on the input RGB image. For the fusion-based XRF image inpainting algorithm, the XRF spectrum of each pixel is represented by an endmember dictionary, as well as the RGB spectrum. The input subsampled XRF image is decomposed into visible and non-visible components, making it possible to find the nonlinear mapping from the RGB to the XRF spectrum. Experiments show the effectiveness of our proposed network $NetM$ in both RGB and XRF image sampling and reconstruction tasks. Higher reconstruction accuracy is achieved in both RGB and XRF image sampling, and also the computation time of mask generation is significantly smaller than other adaptive sampling methods.

\bibliographystyle{IEEEtran}
%\newpage
\bibliography{ref}

%\section{Optimization scheme for Equation~\ref{e:e26}}


% \end{IEEEbiography}

%\newpage



%\vfill


\end{document}